\documentclass{article}

\usepackage{natbib}
\setcitestyle{numbers,square}

\usepackage[preprint]{neurips_2023}

\usepackage[utf8]{inputenc} 
\usepackage[T1]{fontenc}    
\usepackage{hyperref}       
\usepackage{url}            
\usepackage{booktabs}       
\usepackage{amsfonts}       
\usepackage{nicefrac}       
\usepackage{microtype}      
\usepackage{xcolor}         
\usepackage{ulem}
\usepackage{graphicx}
\usepackage{subfigure}
\usepackage{amsmath}
\usepackage{multirow}
\usepackage{colortbl}
\usepackage{wrapfig}
\usepackage{hyperref}

\title{Unifying gradient regularization for Heterogeneous Graph Neural Networks}

\author{%
  Xiao Yang, Xuejiao Zhao\protect\footnotemark[1], Zhiqi Shen\protect\footnotemark[1] \\
 School of Computer Science and Engineering\\
  Nanyang Technological University\\
  Singapore \\
}

\begin{document}

\maketitle

\renewcommand{\thefootnote}{\fnsymbol{footnote}}
\footnotetext[1]{Corresponding Author}
\renewcommand{\thefootnote}{\arabic{footnote}}

\begin{abstract}
Heterogeneous Graph Neural Networks (HGNNs) are a class of powerful deep learning methods widely used to learn representations of heterogeneous graphs. Despite the fast development of HGNNs, they still face some challenges such as \textit{over-smoothing}, and \textit{non-robustness}. Previous studies have shown that these problems can be reduced by using gradient regularization methods. However, the existing gradient regularization methods focus on either graph topology or node features. There is no universal approach to integrate these features, which severely affects the efficiency of regularization.
In addition, the inclusion of gradient regularization into HGNNs sometimes leads to some problems, such as an unstable training process, increased complexity and insufficient coverage of regularized information.
Furthermore, there is still short of a complete theoretical analysis of the effects of gradient regularization on HGNNs. In this paper, we propose a novel gradient regularization method called \textit{Grug}, which iteratively applies regularization to the gradients generated by both propagated messages and the node features during the message-passing process. \textit{Grug} provides a unified framework integrating graph topology and node features, based on which we conduct a detailed theoretical analysis of their effectiveness. Specifically, the theoretical analyses elaborate the advantages of \textit{Grug}: 1) Decreasing sample variance during the training process~\textbf{(Stability)}; 2) Enhancing the generalization of the model~\textbf{(Universality)}; 3) Reducing the complexity of the model~\textbf{(Simplicity)}; 4) Improving the integrity and diversity of graph information utilization~\textbf{(Diversity)}. As a result, \textit{Grug} has the potential to surpass the theoretical upper bounds set by DropMessage.~\footnote{AAAI-23 Distinguished Papers}. In addition, we evaluate \textit{Grug} on five public real-world datasets with two downstream tasks. The experimental results show that \textit{Grug} significantly improves performance and effectiveness, and reduces the challenges mentioned above. Our code is available at: \href{https://github.com/YX-cloud/Grug}https://github.com/YX-cloud/Grug.
\end{abstract}

\section{Introduction}
As graph neural networks (GNNs) \cite{battaglia2018relational,kipf2016semi} have emerged as powerful architectures for learning and analyzing graph representations. Therefore, in recent years, researchers have actively explored their potential in Heterogeneous Graph Neural Networks (HGNNs)~\cite{dong2017metapath2vec,schlichtkrull2018modeling,wang2020relational}. Heterogeneous graphs consist of multiple types of nodes and edges with different side information. To tackle the challenge of heterogeneity, various HGNNs\cite{quadrana2018sequence,tay2018multi} have been proposed, such as HGT\cite{hu2020heterogeneous}, HAN\cite{wang2019heterogeneous}, fastGTN\cite{yun2022graph}, SimpleHGNN\cite{lv2021we}, MAGNN\cite{fu2020magnn}, RSHN\cite{zhu2019relation} and HetSANN\cite{hong2020attention} were developed within recent years. These HGNNs perform well in a variety of downstream tasks, including node classification, link prediction, and recommendation \cite{zhang2019heterogeneous,zhao2022heterogeneous}.

Despite the rapid development of the HGNNs, they often face problems such as \textit{over-smoothing} and \textit{non-robustness}~\cite{wang2022ensemble}, which seriously affects the efficiency of downstream tasks. Heterogeneous graphs (HGs) are more ubiquitous than homogeneous graphs in real-world scenarios, but they are more difficult to represent than homogeneous graphs. Because the diversity of data structures of HGs limits the generalization of HGNNs and reduces their robustness as well. Moreover, during message-passing in HGNNs, each node aggregates messages from its neighbors in each layer, HGNNs tend to blur the features of nodes, which is called over-smoothing. Some studies have proved that HGNNs are a special kind of regularization, and their \textit{over-smoothing} and \textit{non-robustness} problems mentioned above can be effectively alleviated by using additional regularization to the HGNNs~\cite{wang2022ensemble,zhu2021interpreting}.

Gradient regularization is one of the most popular regularization methods to improve the above problem. It integrates various regularizations into one gradient through different strategies. Many widely used regularization methods have greatly improved the performance of HGNNs, such as random dropping methods~\cite{hinton2012improving,rong2019dropedge,feng2020graph,fang2022dropmessage}, Laplacian regularization methods~\cite{you2020l2, yang2021rethinking} and adversarial training methods~\cite{xu2022graph,jin2020graph,wu2020graph,kong2020flag}, etc.
In this paper, we prove theoretically that these methods are essentially a special form of gradient regularization.

Although gradient regularization methods have been widely applied to HGNNs, there are still some open questions that need to be addressed. First, the existing gradient regularization methods only perform regularization on one of the following three information dimensions, namely node, edge and propagation message. These methods do not fully explore which dimension is the optimal solution. Second, the inclusion of gradient regularization into HGNNs sometimes leads to some additional problems, such as unstable training process, and parameter convergence difficulty caused by multiple solutions problems. Lastly, there is still short of a complete theoretical analysis of the effects of gradient regularization on HGNNs.

In this paper, we propose a novel gradient regularization method called \textit{Grug}, which can be applied to all message-passing HGNNs. \textit{Grug} provides a unified framework integrating graph topology and node features by iteratively applying regularization to the gradients generated by both propagated messages and the node features during the message-passing process. We conduct a detailed theoretical analysis of the effectiveness of \textit{Grug}. These theoretical analyses elaborate the advantages of \textit{Grug} in 4 dimensions: 1) \textit{Grug} makes the training process more stable by realizing the controllable variance of the input samples~\textbf{(Stability)}; 2) \textit{Grug} unify a wider range of regularizations to improve the performance of HGNNs~\textbf{(Universality)}; 3) \textit{Grug} ensures the continuity of gradient and the unique solution of the model during gradient regularization, which alleviates parameter convergence difficulty caused by traditional multiple solutions models~\textbf{(Simplicity)}; 4) \textit{Grug} can provide more diverse information to the model and improve the integrity and diversity of graph information utilization~\textbf{(Diversity)}. Thus, \textit{Grug} has the potential to surpass the DropMessage~\cite{fang2022dropmessage}. In addition, we evaluate \textit{Grug} on five public real-world datasets with two downstream tasks. Our main contributions are summarized as follows:

\begin{itemize}
\item We propose a novel gradient regularization method for all message-passing HGNNs named \textit{Grug}. \textit{Grug} can unify the existing gradient regularizations on HGNNs into one framework by resetting distribution (dropping or perturbation) in certain rules on the feature matrix and message matrix. In other words, the existing methods can be regarded as one special form of \textit{Grug}.

\item We provide a series of comprehensive and detailed theoretical analyses covering the stability, universality, simplicity and diversity of gradient regularization methods. Which sufficiently explains "Why" and "How" \textit{Grug} has
the potential to surpass the theoretical upper bounds set by DropMessage~\cite{fang2022dropmessage}.

\item We conduct sufficient experiments for different downstream tasks and the results show that \textit{Grug} not only surpasses the state-of-the-art~(SOTA) performance, but also alleviates over-smoothing, and non-robustness obviously compared to the existing gradient regularization methods.

\end{itemize}

\section{Related Work}~\label{RelatedWork}
In general, random dropping methods and adversarial training methods can be regarded as a form of gradient regularization methods~\cite{bishop1995training,simon2019first,qin2019adversarial}. As a representative of random dropping, Dropout \cite{hinton2012improving,burges1996improving} has been proven to be effective in many fields~\cite{maaten2013learning} with enough theoretical analysis~\cite{wager2013dropout}. Besides, applying random dropping to HGNNs can achieve effective performance, such as DropNode~\cite{feng2020graph} and DropEdge\cite{rong2019dropedge}, which randomly drop nodes and edges during the training process respectively. Additionally, the best random dropping method is DropMessage~\cite{fang2022dropmessage}, which applies random dropping on propagated messages. Adversarial Training is proposed as a method of defense against attacks~\cite{goodfellow2014explaining}, which generate new adversarial samples according to the gradient and adds them to the training set. On this basis, a stronger adversarial attack model PGD~\cite{madry2017towards} has become the mainstream model. In fact, adversarial training has been shown to be closely related to regularization~\cite{simon2019first,qin2019adversarial}. Recently adversarial training has also been applied to HGNNs~\cite{xu2022graph,jin2020graph,wu2020graph,zhang2020gnnguard}. Specially, FLAG~\cite{kong2020flag} tries to establish node-level adversarial training to improve the representation of the graph and achieve satisfactory results. These existing gradient regularization methods can be used to relieve over-smoothing and over-fitting problems~\cite{fang2022dropmessage,kong2020flag,feng2020graph}. But all of them only perform regularization on one of the following three information dimensions, namely node, edge and propagation message. They do not fully explore which dimension is the optimal solution~\cite{xu2022graph,wu2020graph,zhang2020gnnguard}. Furthermore, we find that the inclusion of gradient regularization into HGNNs sometimes leads to some additional problems, such as unstable training process, and parameter convergence difficulty caused by multiple solutions problems. All the problems of the existing methods are alleviated by the proposed method in this paper.

\section{Notations and Preliminaries}
\textbf{Notations.} For simplicity of analysis, we assume that the heterogeneous graph contains 2 relations, $r$ and $r^\sim$. \textbf{HG} = (\textbf{V},\textbf{E},\textbf{R}) represent the heterogeneous graph, where \textbf{V} = \{$v_1 ... v_p$\} denotes $p$ heterogeneous nodes, and \textbf{E}  = \{$e_1 ... e_q$\} denotes $q$ heterogeneous edges. \textbf{R} = \{$r$ , $r^\sim$\} represents relations in heterogeneous graph. The node features can be denoted as a matrix $F = \{x_1 ... x_n, x_p\} \in R^{p \times d}$, where $x_i$ is the feature vector of the node $v_i$, and $d$ is the dimension of node features. The edges and relations describe the relations between nodes and can be denoted as an adjacent matrix $A = \{ a_1 ... a_q \} \subseteq R^{q \times q }$, where $a_i$ denotes the $i$-th row of the adjacent matrix, and $A(i, j)$ denotes the relation between node $v_i$ and node $v_j$. When we apply message-passing HGNNs on \textbf{HG}, the message matrix can be represented as $M = \{ m_1 ... m_k \} \subseteq R^{k \times d'}$, where $m_i$ is a message propagated between nodes, $k$ is the number of messages propagated on the heterogeneous graph and $d'$ is the dimension of the messages.

\textbf{Message-passing HGNNs.} 
In most existing HGNN models, a message-passing framework is adopted, where each node sends and receives messages to/from its neighbors based on their relations. During the propagation process, node representations are updated using both the original node features and the received messages from neighbors. This process can be expressed as follows:

\begin{equation}
    \begin{aligned}
         h_i^{(l+1)}&= \sigma^{(l)} (h_i^{(l)}, AGG_{j \in N(i)} {\phi^{(l)}(h_i^{(l)}, h_j^{(l)}, e_{j,i})})
    \end{aligned}
\end{equation}

where $h_i^{(l+1)}$ denotes the representation of node $v_i$ in the $l$-th layer, and $N(i)$ is a set of adjacent nodes to node $v_i$; $e_{j,i}$ represents the relation from node $v_i$ to node $v_j$; $AGG$ denotes the aggregation operation; and $\sigma^{(l)}$ and $\phi^{(l)}$ are differentiable functions. Besides we can gather all propagated messages into a message matrix $M \subseteq R^{k \times d'}$, which can be expressed as below:

$$M_i^{(l+1)} =  {\phi(h_i^{(l)}, h_j^{(l)}, e_{j,i})}$$

where $\phi$ denotes the message mapping information, and the row number $k$ of the message matrix M is equal to the directed edge number in the graph.

\section{Our Approach}

\begin{figure}[htbp]
\vspace{-1.0cm}
    \centering
    \includegraphics[scale=0.5]{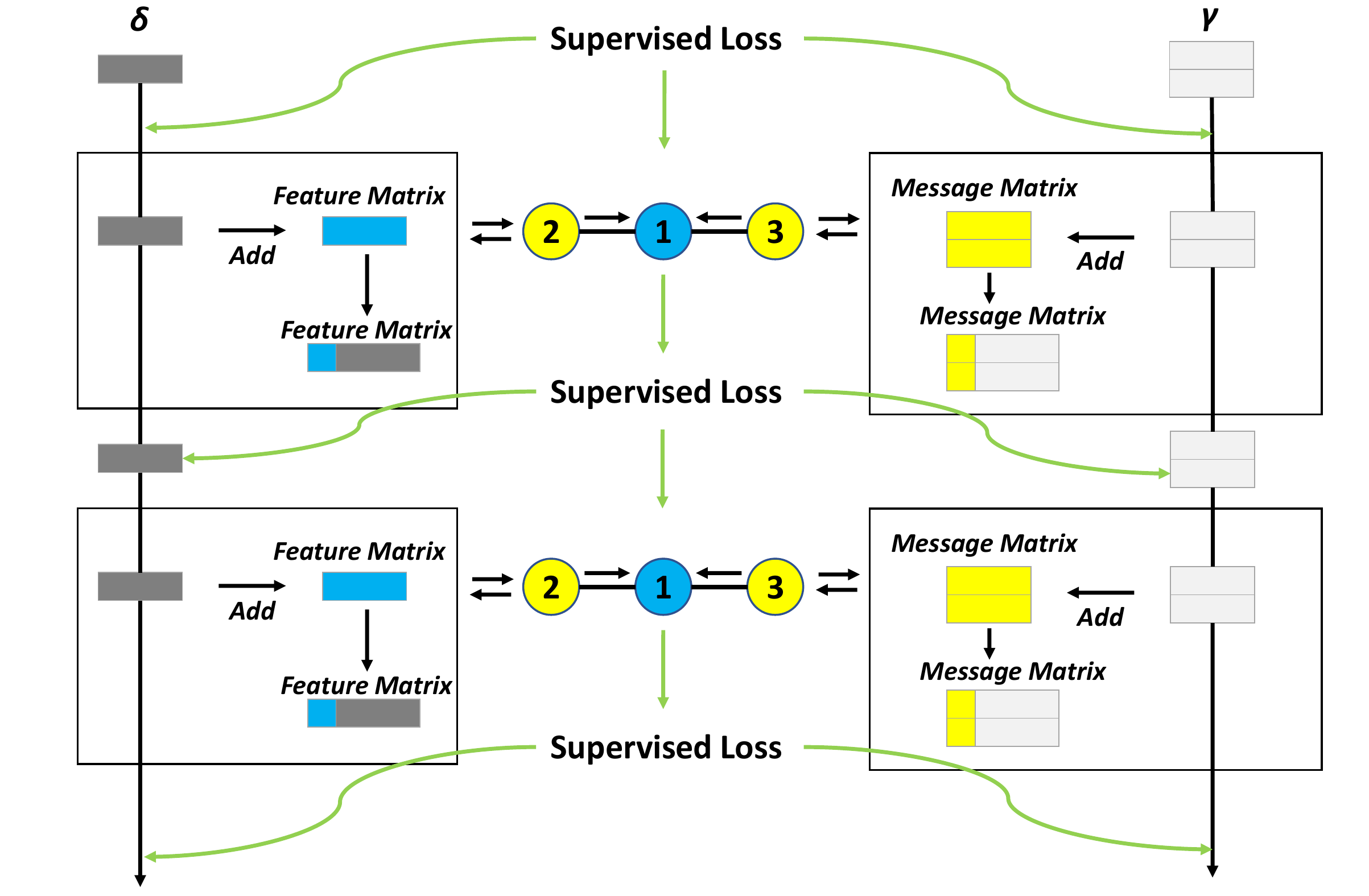}
    \caption{Illustrations of \textit{Grug}. Considering the messages propagated by
the center node (i.e., Node 1), \textit{Grug} allows to perturb the feature matrix and message matrix, which implements the regularization of gradients on the feature matrix and message matrix.}
\label{framework}
\vspace{-0.3cm}
    \end{figure}

In this section, we introduce our newly proposed method called \textit{Grug}, which can be applied to all message-passing HGNNs. We first introduce details of our \textit{Grug}, and further prove that the most common gradient regularization methods in HGNNs such as random dropping and adversarial perturbations, can be unified into our framework. Then, we present the theoretical evidence of the effectiveness of these methods. Furthermore, we analyze the superiority of \textit{Grug}, and provide a series of theoretical proofs from the aspects of stability, universality, simplicity and diversity.

\subsection{\textit{Grug}}
\label{Algorithm description}
\textbf{Algorithm description.} 

We apply a single RGCN as the backbone model, which can be formulated as $f(F, M, W) = FMW$, where $F \in R^{n\times k}$ denotes the feature matrix, $M$ indicates the message matrix and $W$ denotes the transformation matrix. When we use cross-entropy as the loss function, the objective function can be expressed as follows:
\begin{equation} \label{loss function}
    \begin{aligned}
        L(f(F,M,W),y) &= \sum_{i=0}^{K-1} y_i \log{(softmax(f(F,M,W)))}
    \end{aligned}
\end{equation}
where $K$ is the class number and $y_i$ is one hot representation of label $i$.

Different from existing gradient regularization methods, \textit{Grug} performs gradient regularization both on the feature matrix $F$ and message matrix $M$ instead of either one of them and the process of \textit{Grug} is in Figure \ref{framework}.
For each element $F_{i,j}$ in $F$ and $M_{i,j}$ in $M$, we generate trained additional elements $\gamma^{i, j}$ and $\delta ^{i, j}$ to perturb the original distribution, according to uniform distribution $\gamma ^{i, j} \sim  Uniform(-\alpha,\alpha)$ and $\delta^{i, j} \sim  Uniform(-\beta,\beta)$. After that, we obtain the perturbed $F$ and $M$ by adding each element with its trained additional element. Besides, in the training process, $\gamma ^{i, j}$ and $\delta^{i, j}$ can be updated according to the gradient of the previous epoch, which can be expressed as follows:

\begin{equation}
    \begin{aligned}
         \delta_t &=\delta_{t-1} + \beta \cdot 
 \frac{\partial_{\delta} L(f(F,M,W),y),y)}{||\partial_{\delta} L(f(F,M,W),y),y)||}\\
    \end{aligned}
\end{equation}
 

\textbf{Unifying gradient regularization methods.} \label{unify}

In this section, we show that existing gradient regularization methods can be integrated into our \textit{Grug}. Different from these methods, \textit{Grug} performs regularization rules on the feature matrix and message matrix. First, we demonstrate that Dropout~\cite{wager2013dropout}, DropEdge~\cite{rong2019dropedge}, DropNode~\cite{feng2020graph}, DropMessage~\cite{fang2022dropmessage} and FLAG~\cite{kong2020flag} can all be formulated as norm regularization process with specific dimension. More importantly, we find that existing methods are a special case of \textit{Grug} in terms of  norm dimensions. In summary, \textit{Grug} can be expressed as a unified framework of gradient regularization methods on HGNNs.

\textbf{Theorem 1.} \textit{Dropout, DropEdge, DropNode, DropMessage and FLAG can be formulated in specific-dimension norm regularization process, and \textit{Grug} is the generalization case of these methods.} \label{Theorem 1}
We show the regularization term of each method below.

\textit{Dropout.}~dropping the elements $F_{i,j} = \eta F_{i,j}$ where $\eta \sim Bernoulli(\mu)$ is equivalent to perform regularization term $\frac{1}{2} \cdot || \partial_{F_\zeta} {L} ||_0$ on message matrix.

\textit{DropEdge.}~dropping the elements $A_{i,j} = \epsilon A_{i,j}$ where $\epsilon \sim Bernoulli(\mu)$ is  equivalent to perform regularization term $\frac{ \partial_A M}{2} \cdot || \partial_{M _ \epsilon} {L} ||_0$ on message matrix.

\textit{DropNode.}~dropping the elements $F_i = \eta F_i$ where $\zeta \sim Bernoulli(\mu)$ is equivalent to perform regularization term $frac{1}{2} \cdot || \partial_{F_\zeta} {L} ||_0$ on feature matrix.

\textit{DropMessage.}~dropping the elements $M_{i,j}= \tau M_{i,j}$ where $\tau \sim Bernoulli(\mu)$ is equivalent to perform regularization term $\frac{1}{2} \cdot || \partial_{M _ \tau} {L} ||_0$ on message matrix.

\textit{FLAG.}~Perturbing the feature matrix $F+\delta$, which is equivalent to perform regularization term $\varepsilon_t || \partial_F L ||_{q_t}$ on feature matrix. 

\textit{Grug.}~Perturbing both the feature matrix $F+\delta$ and message matrix $M+\xi$, which is equivalent to perform regularization term $\varepsilon_t \cdot || \partial_F L ||_{q_t} + \xi_t  \cdot || \partial_F L ||_{h_t}$ on feature matrix and message matrix. Where $\delta$ and $\gamma$ are the trained $F$ perturbation, which are defined as follows:

\begin{equation}
    \begin{aligned}
                \delta_t &=\delta_{t-1} + \beta \cdot 
 \frac{\partial_{\delta} L(f(F+\delta_{t-1},M,W),y)}{||\partial_{\delta} L(f(F+\delta_{t-1},M,W),y)||}\\
                \gamma_t &=\gamma_{t-1} + \alpha \cdot 
 \frac{\partial_{\alpha} L(f(F+\gamma_{t-1},M,W),y)}{||\partial_{\gamma} L(f(F+\gamma_{t-1},M,W),y)||}
    \end{aligned}
\end{equation}

 $||\cdot||_{p_t}$ and $||\cdot||_{l_t}$ are the dual norm of $||\cdot||_{q_t}$ and $||\cdot||_{h_t}$, which are defined as follows:
\begin{equation}
    \begin{aligned}
        ||z||_q &= sup\{z^Tx \enspace  ||x||_p \le 1\} \enspace and  \enspace \frac{1}{p} + \frac{1}{q}  =1\\
        ||z||_h &= sup\{z^Tx \enspace  ||x||_l \le 1\} \enspace and  \enspace \frac{1}{h} + \frac{1}{l}  =1
    \end{aligned}
\end{equation}

We find that the range of regularization dimension of \textit{Grug} is from $0$ to $\infty$. While for \textit{Dropout}, \textit{DropEdge}, \textit{DropNode} and \textit{DropMessage} the range of regularization dimension is 0. It is demonstrated that these four methods are special cases of $Grug$. Compared to $FLAG$, $Grug$ can perform different-dimension norm regularization on the feature matrix and message matrix respectively instead of maintaining a consistent-dimension. Thorough proof can be found in Appendix \ref{proof}.

\subsection{Advantage of \textit{Grug}.} \label{advantage}
In this section, we present four further theoretical analyses to elaborate the benefits of \textit{Grug} from the aspects of stability (decreasing sample variance during the training process), universality(), simplicity (reducing complexity of the model), diversity (improving the integrity and diversity of graph information utilization).

\textbf{1) Stability.} \label{stability}

Existing gradient regularization methods face the problem of sample stability, which is caused by the data noise introduced by different operations, such as dropping and perturbation. Data noise brings the problem of instability in the training process. Generally, sample variance $V$ can be used to
measure the degree of stability. A small sample variance means that the training process will be more stable.

\textbf{Theorem 2.} \textit{Grug can adjust the sample variance to achieve the stability of the training process.}

Intuitively, \textit{Grug} achieves regularization by adding uniform distribution to the input samples rather than dropping graph data, which means that the sample variance of the input in each epoch is irrelevant to the graph data. In other words, \textit{Grug} can tune hyperparameters $\alpha$ and $\beta$ to limit the range of sample variance. By reducing the sample variance, \textit{Grug} diminishes the difference between the feature matrix and message matrix, which stabilizes the training process. Specific range inference can be found in Appendix \ref{proof}.

\textbf{2) Universality.} \label{universality}

\textbf{Theorem 3.} \textit{Grug performs regularization on the feature matrix and message matrix, which is the optimal general solution compared to existing regularization methods.}

As we summarized in Section~\ref{RelatedWork}, all existing gradient regularization methods only perform regularization on one of the following three information dimensions, namely node, edge and propagation message. In order to explore which dimension is the optimal solution, we first need to know the relations between these three information dimensions. The relations between these three information dimensions can be expressed as follows:

{Regularization on message dimension.}
 $$ \partial_M L = \partial_F L  \cdot  \partial_M F \quad and \quad \partial_M L = \partial_A L  \cdot  \partial_M A
$$ 

{Regularization on edge dimension.}
 $$ \partial_A L = \partial_M L  \cdot  \partial_A M \quad and \quad\partial_A L = \partial_F L \cdot \partial_M F \cdot  \partial_A M
$$

{Regularization on node dimension.}
 $$ \partial_F L  = \partial_M L  \cdot  \partial_F M \quad and \quad \partial_F L  = \partial_A L  \cdot  \partial_M A  \cdot  \partial_F M$$ 

The above expressions show that the gradient regularization on each information dimension is transferred to the other two information dimensions. Therefore, regularization on any one information dimension is equivalent to regularization on all three information dimensions theoretically. However, we find that these transfers are not in real-time. Take regularization on the node as an example.
In epoch $e_i$, the gradient of $F$ can be calculated by $L$, which we express as $\partial_F L_i$. There is no gradient of $F$ generated from $M$, because the model only keeps the gradient from the loss. However, $\partial_F M$ must exist objectively, because $M$ can be seen as a function involving $F$ and $A$. So we can infer that in epoch $e_i$, $\partial_F M$ is correlated to the previous $e_{i-1}$ epoch. 
This non-real-time gradient transfer phenomenon is also the same in the edge dimension and message dimension. So we conclude that regularization on only one information dimension cannot apply regularization in real-time to other dimensions. So, gradient regularization on all three information dimensions simultaneously is a better strategy than gradient regularization on one information dimension.

But in practice, we find that compared with node and message dimensions, performing gradient regularization on edge dimension is not only memory consumption, but also reduces the training seed sharply. On the other hand, according to the loss function~\ref{loss function}, the node dimension and message dimension, rather than the edge dimension, have the greatest impact on the model loss. Furthermore, performing regularization on the node dimension and message dimension also includes the edge dimension partially. The results of additional experiments show that there are no significant differences between the effects of performing gradient regularization on the edge dimension or not. Therefore, \textit{Grug} performing regularization on the message and node dimension is the optimal strategy. The Universality is the key for \textit{Grug} to surpass theoretical upper bounds set by DropMessage and detailed proof are presented in Appendix \ref{proof}.

\textbf{3) Simplicity.} \label{simplicity}

\textbf{Theorem 4.} \textit{Message-passing HGNNs using Grug remain one and unique solution, which maintains parameters easily training.}

According to Theorem 1 in section \ref{Theorem 1}, models using dropping operations on information dimensions can be uniformly expressed as:

\begin{equation}
    \begin{aligned}
Drop: \quad \min {||\partial_X L||_0} \quad and \quad X = F ,  M , A\\
Gurg: \quad \min {||\partial_M L||_{q_t} + ||\partial_F L||_{h_t}}
    \end{aligned}
\end{equation}

Here $||\cdot||_0$ represents the 0-dimension Paradigm. However,  0-dimension Paradigm is NP-hardness (non-deterministic polynomial-time hardness)~\cite{donoho2006most} and non-convex~\cite{nguyen2019np}. Thus, it is hard to find the solution of functions related to dropping, which makes it hard to run at a satisfactory speed.

In formula (6), the $q_t$ and $h_t$ are hyperparameters and we can control them by $\alpha$ and $\beta$.  So our $Grug$ uses the 1-dimension Paradigm or 2-dimension Paradigm. In theory, the 1-dimension Paradigm \cite{donoho2006most} and the 2-dimension \cite{ke2003robust} Paradigm are all convex and remain one and unique solution. Compared to Dropout, DropNode, DropEdge and DropMessage, \textit{Grug} will not escalate the difficulty of training from the perspective of convexity. The detailed proof is exhibited in Appendix \ref{proof}.

\textbf{4) Diversity.} \label{diversity}

\textbf{Theorem 5.} \textit{Grug is able to maximize the utilization of diverse information in heterogeneous graph.}

Utilization of diverse information contains 2 aspects: 1)~Obtaining more diverse information 2)~Keeping original information diversity. 

Obtaining more diverse information is defined as the total number of sample distributions generated during the training process. We propose that the total number of elements in the message matrix is $Z$. For dropping methods, the number of new data distributions is generated by DropMessage, which can be expressed as $C_{Z} ^{\frac{Z}{2}}$ and $C$ is a combination in mathematics. Our \textit{Grug} can continue to generate new data distribution until the end of training.

keeping original information diversity is defined as the total number of preserved feature dimensions. In this term, DropMessage has been proven to perform best in dropping operations. However, \textit{Grug} seems to lose initial information rarely. This is because perturbation $\delta$ and $\gamma$ will gradually reduce the value of each element as the training progresses. So \textit{Grug} may lose information only in the initial stage of training.

From the perspective of these 2 aspects, \textit{Grug} performs better than existing methods. Although \textit{Grug} may lose some information as other methods, it generates immense new sample distributions, thus maximizing the utilization of diverse information. More details of the derivation are shown in Appendix~\ref{proof} and supplementary experiments can be found in Appendix~\ref{Additional Experiments}.

\section{Experiments}

\subsection{Experimental Setup}
In this section, we present the results of our experiments on five real-world datasets for node classification and link prediction tasks. Our experiments aim to answer the following questions: 1) How does \textit{Grug} compare to other gradient regularization methods on HGNNs in terms of performance? 2) Can \textit{Grug} improve the robustness of HGNNs? 3) Does \textit{Grug} perform better than other methods in addressing over-smoothing problems? 

\textbf{Datasets} We employ five commonly used heterogeneous graph datasets in our experiments, which are \textit{ACM}, \textit{DBLP}, \textit{IMDB}, \textit{Amazon} and \textit{LastFM}.

\textit{Node Classification Datasets.} Labels in node classification datasets are split according to 20\% for training, 10\% for validation and
70\% for test in each dataset. Note that node classification datasets are the same version used in the GTN~\cite{yun2019graph}.
\begin{itemize}
    \item \textbf{ACM} is a paper citation network, containing 3 types of nodes~(papers~(P), authors~(A), subjects~(S)), 4 types of edges~(PS, SP, PA, AP).
    \item \textbf{DBLP} is a bibliography website of computer science. It contains 3 types of nodes~(papers (P), authors (A), conferences (C)), 4 types of edges~(PA, AP, PC, CP).
    \item \textbf{IMDB} is a website about movies and related information. It contains 3 types of nodes (movies (M), actors (A), and directors (D)) and labels are genres of movies. 
\end{itemize} 

\textit{Link Prediction Datasets.} Edges in link prediction datasets are split according to 25\% for training, 5\% for validation and
60\% for the test in each dataset. Note that link prediction datasets are subsets pre-processed by HGB dataset~\cite{lv2021we}.
\begin{itemize}
    \item \textbf{Amazon} is an online purchasing platform and it consists of electronics products with co-viewing and co-purchasing links between them.
    \item \textbf{LastFM} is an online music website. It contains 3 types of nodes~(artist, user, tag) and 3 types of edges~(artist-tag, user-artist, user-user).
\end{itemize}
 \textbf{Baseline methods.} We compare our proposed \textit{Grug} with other existing gradient regularization methods, including Dropout (2012)~\cite{hinton2012improving}, DropNode (2020)~\cite{feng2020graph}, DropMessage (2022)~\cite{fang2022dropmessage}, and FLAG (2020)~\cite{kong2020flag}. We adopt these methods on backbone models and compare their performances on different datasets.

 \textbf{Backbone models.} In order to compare the differences of gradient regularization methods fairly, we choose the basic HGNN models to ensure that the experimental results are not affected by the difference of the models. So we consider two mainstream HGNNs as our backbone models: RGCN~\cite{schlichtkrull2018modeling} and RGAT~\cite{wang2020relational}. We implement the gradient regularization methods mentioned above on backbone models.

\begin{table}[htbp]
\centering
\scalebox{0.85}{\begin{tabular}{ccc|cccccc} 
\hline
\multicolumn{3}{c}{Dataset}                            & \multicolumn{2}{c}{ACM}                                                                                             & \multicolumn{2}{c}{DBLP}                                                                                            & \multicolumn{2}{c}{IMDB}                                                                                             \\ 
\hline
\multicolumn{3}{c}{Method}         & Macro F1                                     & Macro F1                                                & Micro F1                                                & Macro F1                                                & Micro F1                                                & Macro F1                                                 \\ 
\hline
\multirow{6}{*}{\rotatebox{90}{\textbf{RGCN}}} & \multicolumn{2}{c}{Clean \cite{schlichtkrull2018modeling}}    
& 90.36±{\small 0.60 }                                   
& 90.44±{\small 0.59 }                                             & 93.67±{\small 0.15}                                              & 92.77±{\small 0.16  }                                            & 56.05±{\small 1.84 }                                             & 55.43±{\small 1.64}                                             \\
                               & \multicolumn{2}{c}{Dropout~\cite{hinton2012improving}}       
                                & 90.86±{\small 0.39}
                               & 90.91±{\small 0.40}                                             & 94.35±{\small 0.14}                                              & 93.53±{\small 0.14}                                              & 57.63±{\small 1.01}                                              & 56.04±{\small 0.79}                                               \\
                               & \multicolumn{2}{c}{DropNode~\cite{feng2020graph}} 
                                & 90.68±{\small 0.75}
                               & 90.74±{\small 0.77}                                              & 93.85±{\small 0.71}                                              & 93.12±{\small 0.71}                                              & 57.38±{\small 0.37}                                              & 56.20±{\small 0.50}                                               \\
                               & \multicolumn{2}{c}{DropMessage~\cite{fang2022dropmessage}}   
                                & 91.57±{\small 0.19}   
                               
                               & 91.57±{\small 0.21}                                              & 94.69±{\small 0.13}                                              & 93.96±{\small 0.13}                                              & 57.60±{\small 0.84}                                              & 56.48±{\small 0.49}                                               \\ 
\cline{2-9}
                               & \multicolumn{2}{c}{FLAG~\cite{kong2020flag}}    
                                 & 92.43±{\small 0.37} 
                               
                               & 92.58±{\small 0.34}                                              & 94.22±{\small 0.20}                                              & 93.39±{\small 0.24}                                              & 56.74±{\small 0.33}                                              & 55.63±{\small 0.40}                                               \\
                               & \multicolumn{2}{c}{\textit{Grug}}
                                & \textbf{93.27±{\small 0.13}}
                               
                               & \textbf{93.33±{\small 0.13}} & \textbf{95.08±{\small 0.08}} & \textbf{94.17±{\small 0.11}} & \textbf{59.92±{\small 0.56}} & \textbf{58.87±{\small 0.71}}  \\ 
\hline
\multirow{6}{*}{\rotatebox{90}{\textbf{RGAT}}} & \multicolumn{2}{c}{Clean \cite{schlichtkrull2018modeling}}    
 & 89.32±{\small 1.03}                  

& 89.34±{\small 1.03}                                              & 92.98±{\small 0.61}                                              & 92.15±{\small 0.63}                                              & 53.12±{\small 0.80}                                              & 52.75±{\small 0.78}                                               \\
                               & \multicolumn{2}{c}{Dropout~\cite{hinton2012improving}}     
                               
                                & 89.38±{\small 0.81} 
                               & 89.51±{\small 0.76}                                              & 93.55±{\small 0.48}                                              & 92.69±{\small 0.50}                                              & 54.85±{\small 0.31}                                              & 53.63±{\small 0.26}                                               \\
                               & \multicolumn{2}{c}{DropNode~\cite{feng2020graph}}      
                                  & 89.86±{\small 0.67}  
                               & 89.98±{\small 0.60}                                              & 93.21±{\small 0.67}                                              & 92.42±{\small 0.72}                                              & 53.86±{\small 1.56}                                              & 52.80±{\small 1.60}                                               \\
                               & \multicolumn{2}{c}{DropMessage~\cite{fang2022dropmessage}}
                               & 90.57±{\small 0.62}              
                               
                               & 90.57±{\small 0.62}                                              & 93.89±{\small 0.35}                                              & 93.01±{\small 0.39}                                              & 56.48±{\small 0.76}                                              & 55.29±{\small 0.78}                                               \\ 
\cline{2-9}
                               & \multicolumn{2}{c}{FLAG~\cite{kong2020flag}}   
                                 & 91.46±{\small 0.59}  
                               & 91.53±{\small 0.58}                                              & 93.38±{\small 0.45}                                              & 92.42±{\small 0.35}                                              & 56.09±{\small 0.68}                                              & 54.47±{\small 0.66}                                               \\
                               & \multicolumn{2}{c}{\textbf{Grug}}
                                & \textbf{92.84±{\small 0.18}}
                               & \textbf{92.92±{\small 0.19}} & \textbf{94.21±{\small 0.21}} & \textbf{93.50±{\small 0.22}} & \textbf{57.79±{\small 0.37}} & \textbf{56.17±{\small 0.18}}  \\
\hline
\end{tabular}}
\vspace{2mm}
\caption{Comparison results of different gradient regularization methods for node classification. The best results are in bold.}
\label{table:node}
\vspace{-0.8cm}
\end{table}

\subsection{Comparison Results}
Table \ref{table:link} and Table \ref{table:node} summarize the overall results. For node classification tasks, we employ the Micro F1 score and Macro F1 score as Metrics on 3 datasets (ACM, DBLP, IMDB). For link prediction tasks, the performance is measured by AUC-ROC on Amazon and LastFM. Considering the
space limitation, some additional experiments
are presented in the Appendix \ref{Additional Experiments}.


\begin{wraptable}{i}{7cm}
\centering
\scalebox{0.85}{\begin{tabular}{ccc|cc} 
\hline

\multicolumn{3}{c}{Dataset}                            & Amazon                                                                                            & LastFM                                                                                                                                                                  \\ 
\hline
\multicolumn{3}{c}{Method}         &AUC-ROC                                &AUC-ROC                                                                     \\ 
\hline
\multirow{6}{*}{\rotatebox{90}{\textbf{RGCN}}} & \multicolumn{2}{c}{Clean \cite{schlichtkrull2018modeling}}    
& 70.54±{\small 5.00 }                                   
& 56.66±{\small 5.92 }                                                                                                                      \\
                               & \multicolumn{2}{c}{Dropout~\cite{hinton2012improving}}       
                                & 72.83±{\small 5.17}
                               & 56.95±{\small 2.37}                                                                          \\
                               & \multicolumn{2}{c}{DropNode~\cite{feng2020graph}} 
                                & 70.81±{\small 3.85}
                               & 56.97±{\small 3.86}                                                 \\
                               & \multicolumn{2}{c}{DropMessage~\cite{fang2022dropmessage}}   
                                & \textbf{81.80±{\small 0.57} }  
                               
                               & 58.01±{\small 4.65}                                                \\ 
\cline{2-5}
                               & \multicolumn{2}{c}{FLAG~\cite{kong2020flag}}    
                                 & 78.87±{\small 0.81} 
                               
                               & 56.96±{\small 4.32}                                                    \\
                               & \multicolumn{2}{c}{\textit{Grug}}
                                & 80.93±{\small 0.67}
                               
                               & \textbf{60.37±{\small 1.24}}  \\ 
\hline
\multirow{6}{*}{\rotatebox{90}{\textbf{RGAT}}} & \multicolumn{2}{c}{Clean\cite {schlichtkrull2018modeling}}    
     & 81.98±{\small 2.36} 
                               & 78.09±{\small 3.97}                                                                                         \\
                               & \multicolumn{2}{c}{Dropout~\cite{hinton2012improving}}     
                               
                             & 81.74±{\small 6.72}  
                               & 78.94±{\small 3.43}                                                    \\
                               & \multicolumn{2}{c}{DropNode~\cite{feng2020graph}}      
                                  & 83.36±{\small 0.77}  
                               & 78.10±{\small 2.85}                                                  \\
                               & \multicolumn{2}{c}{DropMessage~\cite{fang2022dropmessage}}
                               & 85.16±{\small 0.35}              
                               
                               & 78.50±{\small 0.52}                                                      \\ 
\cline{2-5}
                               & \multicolumn{2}{c}{FLAG~\cite{kong2020flag}}   
                                 & 85.59±{\small 0.25}  
                               & 78.66±{\small 1.86}                                                   \\
                               & \multicolumn{2}{c}{\textbf{Grug}}
                                & \textbf{86.84±{\small 0.26}}
                               & \textbf{80.39±{\small 2.24}}  \\
\hline
\end{tabular}}
\caption{Comparison results of different gradient regularization methods for link prediction. The best results are in bold.}
\label{table:link}
\vspace{-0.1cm}
\end{wraptable}

Overall, \textit{Grug} performs well in all scenarios (node classification and link prediction), indicating its stability in diverse situations. For node classification tasks, we have 72 settings, each of which is a combination of different backbone models and datasets in different metrics (e.g., RGCN-ACM-Macro F1). Table \ref{table:node} shows that \textit{Grug} achieves the optimal results in all 72 settings. As to 24 settings under the link prediction task, \textit{Grug} achieves the optimal results in 23 settings, and sub-optimal results in 1 setting. Besides, \textit{Grug} performs more stable on all datasets and all tasks than other methods clearly. Taking Dropout as the counterexample, it appears strong performance on the DBLP dataset but shows a clear decrease on the ACM dataset and IMDB dataset. In terms of tasks, the performance of Dropout on link prediction is far worse than that on node classification. Dropout on node classification obtains an average accuracy decline of 2.97\% compared to \textit{Grug}. This numerical gap has widened to 7.69\% on link prediction. A reasonable explanation is that 
the information dimensions operated by distinct methods vary from each other as shown in section \ref{universality}. With the generalization of its operation on node and message dimensions, \textit{DropMessage} obtains a smaller inductive bias. Therefore, \textit{Grug} is more
applicable for most scenarios.

\subsection{Additional Results}

\begin{figure}[!ht]
\vspace{-0.6cm}
\centering
 \subfigure[ACM]{
\includegraphics[scale=0.41]{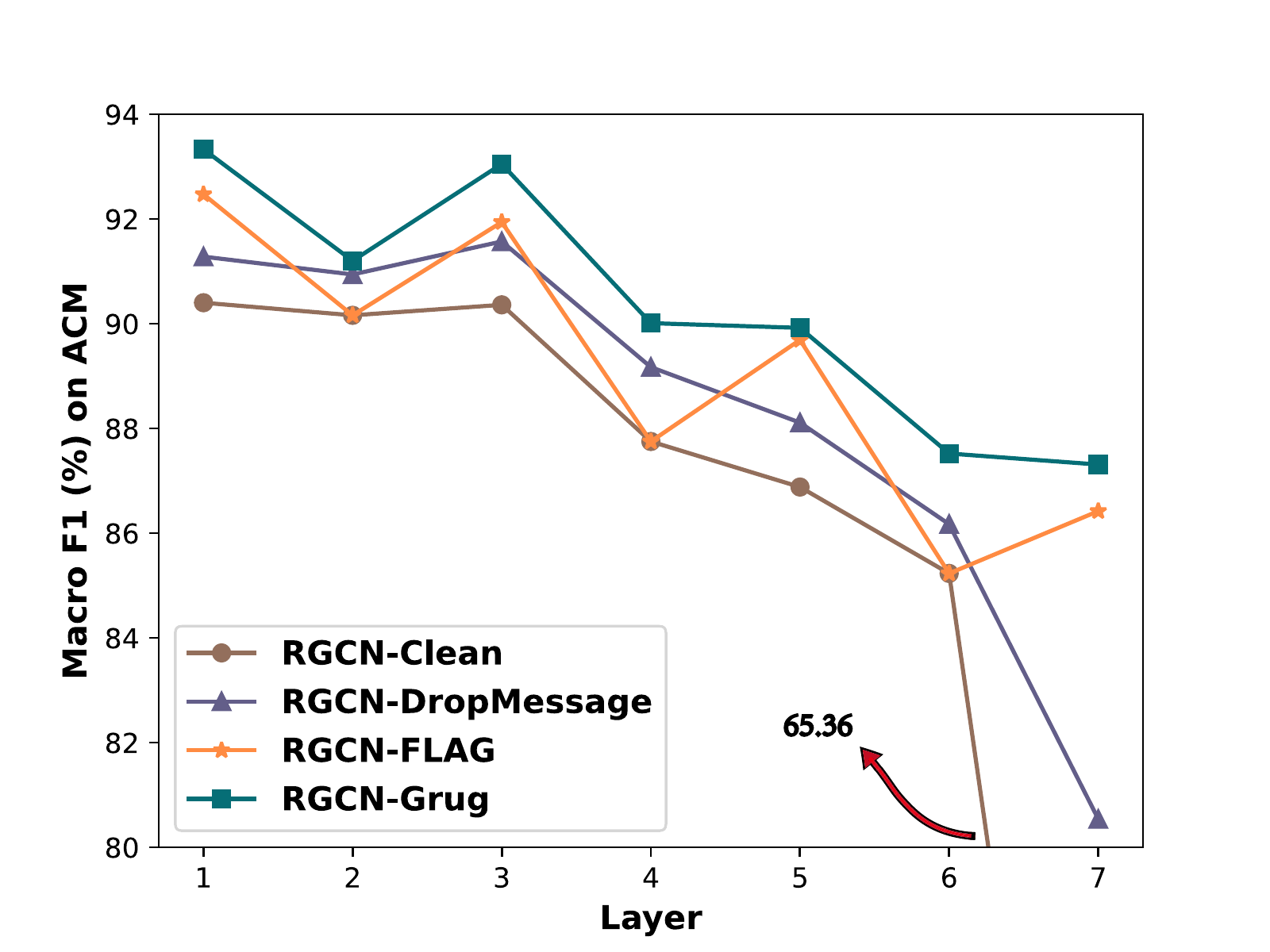}
 \label{fig: Over Smoothing ACM} 
    }
 \subfigure[IMDB]{
\includegraphics[scale=0.41]{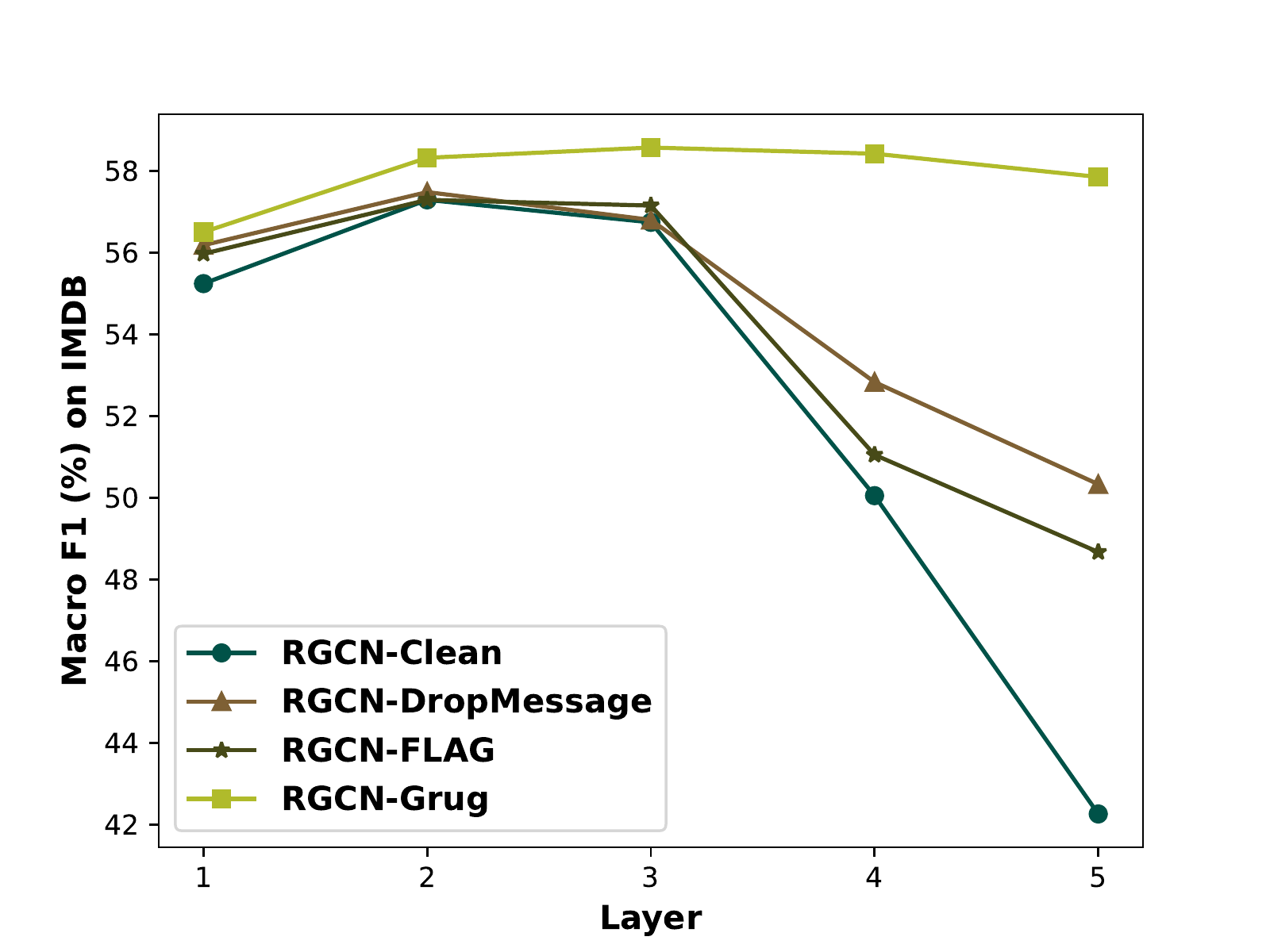}
 \label{fig: Over Smoothing IMDB} 
    }
\caption{Over-Smoothing Analysis.}
\label{fig: Over Smoothing}
\vspace{-0.2cm}
\end{figure}

\textbf{Over Smoothing analysis.} 
In this section, we investigate the issue of over-smoothing in HGNNs, which occurs when node representations become indistinguishable as the depth of the network increases, leading to poor results. To address this problem, we compare the effectiveness of different gradient regularization methods, taking into account the information diversity discussed in Section~\ref{diversity}. As mentioned in section \ref{diversity}, \textit{Grug} obtaining more diverse information in the training process, it is expected to perform better holdout over-smoothing in theory. To verify this, we conduct experiments on two datasets, ACM and IMDB, using RGCN as the backbone model. As Figure \ref{fig: Over Smoothing} suggests, we can find that all gradient regularization methods can alleviate over-smoothing and \textit{Grug}) outperforming other methods due to their diverse information. Our proposed \textit{Grug} exhibits consistent superiority in performance across all layers (from layer 1 to layer 7), as shown in Figure \ref{fig: Over Smoothing}. 

\begin{figure}[!h]
\vspace{-0.6cm}
\centering
 \subfigure[ACM]{
\includegraphics[scale=0.41]{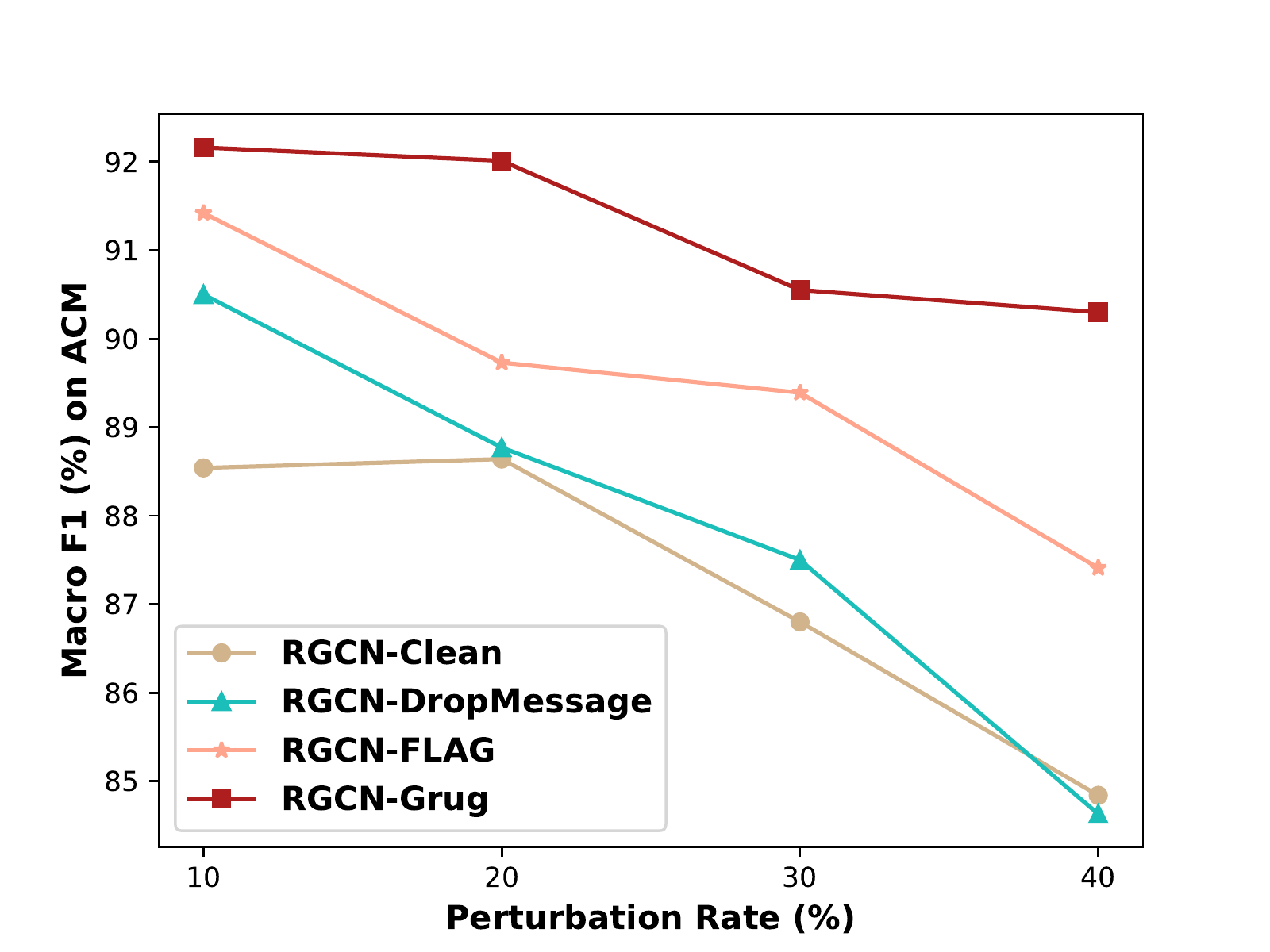}
 \label{fig: robustness ACM} 
    }
 \subfigure[DBLP]{
\includegraphics[scale=0.41]{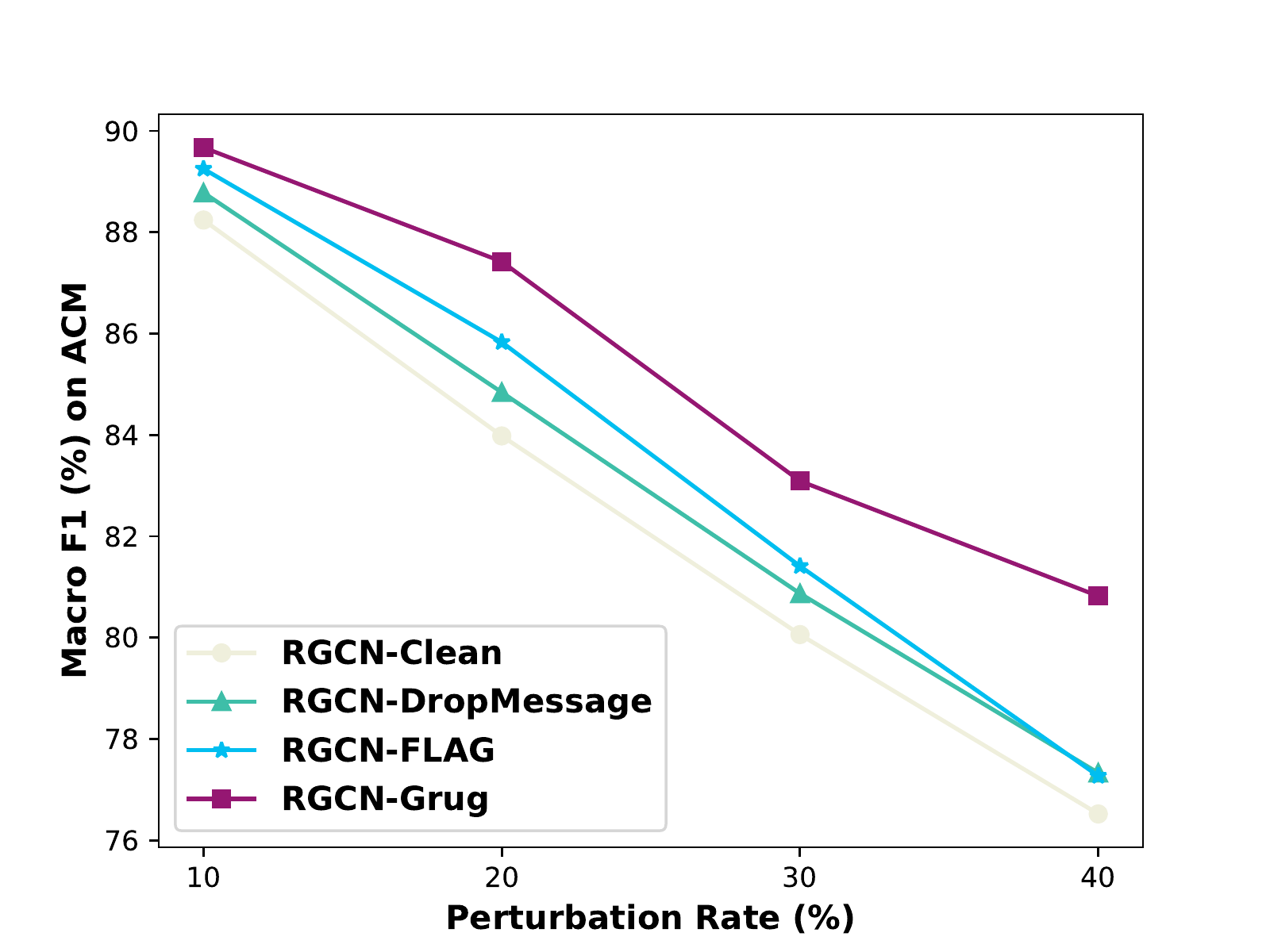}
 \label{fig: robustness DBLP} 
    }
\caption{Model Performances Against Attacks.}
\label{fig: robustness}
\vspace{-0.35cm}
\end{figure}

\textbf{Robustness analysis.} We evaluate the robustness of different gradient regularization methods by measuring their ability to deal with perturbed heterogeneous graphs. In more detail, we randomly add a certain ratio of edges on ACM and DBLP datasets and address the node classification task. It is found in Figure \ref{fig: robustness} that \textit{Grug} has positive effects and results drop the least when the perturbation rate increase from 0.1 to 0.4. Besides, our proposed \textit{Grug} shows its stability and outperforms other gradient regularization methods in noisy situations.

\section{Conclusion}
In this paper, we propose \textit{Grug}, a general and effective gradient regularization method for HGNNs. We first unify existing gradient regularization methods to our framework via performing regularization according to the gradient on the feature matrix and message matrix. As \textit{Grug} operates meticulously on feature and message matrix, it is a generalization of existing methods, which shows greater applicability in general cases. Besides, we provide the theoretical analysis of the superiority~(stability, simplicity and diversity) and effectiveness of \textit{Grug}. By conducting sufficient experiments for multiple tasks on five datasets, we show that \textit{Grug} surpasses the SOTA performance and alleviates over-smoothing, and non-robustness obviously compared to the traditional gradient regularization methods.

\bibliographystyle{unsrt}
\bibliography{neurips_2023}

\clearpage
\appendix
\section{Appendix}
\subsection{Derivation Details.}
\textbf{Detailed proof of Theorem 1.}\label{proof}

\textbf{Theorem 1}~\textit{Dropout, DropEdge, DropNode, DropMessag and FLAG can be formulated specific-dimension norm regularization process and \textit{Grug} is the generalization case of these methods.}

We present more derivation details of Theorem 1. Here we use cross entropy $L$ as the basis loss function, and the objective function $L^\sim$ for each method in expectation can be expressed as follows:

\textit{Dropout:}
\begin{equation}
    \begin{aligned}\notag
        E(L^\sim ) &= E(L^\sim(f(F,M _ \eta ,W),y))\\
        &\approx E(L^\sim(f(F,M _ \eta ,W),y)) + \frac{1}{2} \cdot \partial_M L(f(F,M _ \eta ,W),y)\\
        &\approx L + \frac{1}{2} \cdot \frac{\partial L}{\partial {M _ \eta}} \cdot \frac{\partial{M _ \eta}}{\partial {M}}\\
        & \approx  L+ \frac{1}{2} \cdot || \partial_{M _ \eta} {L} ||_0
    \end{aligned}
\end{equation}
where $\eta \sim Bernoulli(\mu)$ and $M_\eta$ is the dropped $M$. Note that $F_{i,j}= \eta F_{i,j}$ which means that $\mu$ elements of $F$ can be dropped in expectation.

\textit{DropEdge:}
As objective function $L^\sim$ does not explicitly include $A$, so we use the following inference to convert dropping on $A$ to dropping on $M$:
 $$ \partial_A L =  \frac{\partial L}{\partial M} \cdot \frac{\partial M }{\partial A} = \partial_M L  \cdot  \partial_A M
$$    
Therefore, the objective function for DropEdge can be expressed as follows:
\begin{equation}
    \begin{aligned}\notag
        E(L^\sim ) 
        &\approx E(L^\sim(f(F,M _ \epsilon ,W),y)) + \frac{1}{2} \cdot \partial_M L(f(F,M _ \epsilon ,W),y) \cdot \partial_A M \\
        &\approx L + \frac{1}{2} \cdot \frac{\partial L}{\partial {M _ \epsilon}} \cdot \frac{\partial{M _ \epsilon}}{\partial {M}} \cdot  \partial_A M\\
        & \approx  L+ \frac{ \partial_A M}{2} \cdot || \partial_{M _ \epsilon} {L} ||_0
    \end{aligned}
\end{equation}
where $\epsilon \sim Bernoulli(\mu)$ and $M_\epsilon$ is the dropped $M$. Note that $A_{i,j} = \epsilon A_{i,j}$ which means that $\mu$ elements of $A$ and $\mu$ columns of $M$ will be dropped in expectation.

\textit{DropNode:}
\begin{equation}
    \begin{aligned}\notag
        E(L^\sim ) &= E(L^\sim(f(F_\zeta,M,W),y))\\
        &\approx E(L^\sim(f(F_\zeta,M,W),y)) + \frac{1}{2} \cdot \partial_F L(f(F_\zeta,M ,W),y)\\
        &\approx L + \frac{1}{2} \cdot \frac{\partial L}{\partial {F_\zeta}} \cdot \frac{\partial{F_\zeta}}{\partial {F}}\\
        & \approx  L+ \frac{1}{2} \cdot || \partial_{F_\zeta} {L} ||_0
    \end{aligned}
\end{equation}
where $\zeta \sim Bernoulli(\mu)$ and $F_\zeta$ is the dropped $F$. Note that $F_i = \eta F_i$ which means that $\mu$ rows of $F$ will be dropped in expectation.

\textit{DropMessage:}
\begin{equation}
    \begin{aligned}\notag
        E(L^\sim ) &= E(L^\sim(f(F,M _ \tau ,W),y))\\
        &\approx E(L^\sim(f(F,M _ \tau ,W),y)) + \frac{1}{2} \cdot \partial_M L(f(F,M _ \tau ,W),y)\\
        &\approx L + \frac{1}{2} \cdot \frac{\partial L}{\partial {M _ \tau}} \cdot \frac{\partial{M _ \tau}}{\partial {M}}\\
        & \approx  L+ \frac{1}{2} \cdot || \partial_{M _ \tau} {L} ||_0
    \end{aligned}
\end{equation}
where $\tau \sim Bernoulli(\mu)$ and $M_\tau$ is the dropped $M$. Note that $M_{i,j}= \tau M_{i,j}$ which means that $\mu$ elements of $M$ will be dropped in expectation.

\textit{FLAG:}

Note that in each epoch, \textit{FLAG} train parameters $N$ times, and scale the gradient to $\frac{1}{N}$ of the true gradient for accumulation, and update the perturbation $N-1$ times in the process of accumulating the gradient, which is adopted in \textit{Grug}.

\begin{equation}
    \begin{aligned}\notag
        E(L^\sim ) &= \frac{1}{N}\sum_{t=0}^{N-1}{E(L^\sim(f(F+\delta_t,M,W),y))}\\
        &\approx \frac{1}{N}\sum_{t=0}^{N-1} E(L^\sim(f(F,M,W),y))) + \frac{1}{N}\sum_{t=0}^{N-1} \frac{\delta_t}{2} \partial_F L^\sim (f(F,M,W),y)\\
          &= L + \frac{1}{N}\sum_{t=0}^{N-1} \frac{1}{2} \cdot \max_{\delta_t : \left \| \delta_t \right \|_p \le \varepsilon_t}{| L(f(F+\delta_t,M,W),y) - L(f(F,M,W),y) |}\\
           &\approx L + \frac{1}{2N}\sum_{t=0}^{N-1} \cdot \max_{\delta_t : \left \| \delta_t \right \|_p \le \varepsilon_t}{\delta_t \cdot \partial_F L} \\
           &= L + \frac{1}{2N}\sum_{t=0}^{N-1} \varepsilon_t || \partial_F L ||_{q_t}
    \end{aligned}
\end{equation}

where the perturbation $\delta$ iteratively updates N times in each epoch. We can approximate the loss function with the second-order Taylor expansion of $f(\cdot)$ around $\delta$. Thus, the objective loss function in expectation can be expressed as below:
\begin{equation}
    \begin{aligned}\notag
                \delta_t &=\delta_{t-1} + \beta \cdot 
 \frac{\partial_{\delta} L(f(F+\delta_{t-1},M,W),y)}{||\partial_{\delta} L(f(F+\delta_{t-1},M,W),y)||}
    \end{aligned}
\end{equation}
and $||\cdot||_{p_t}$ is the dual norm of $||\cdot||_{q_t}$,which is defined as follows:
\begin{equation}
    \begin{aligned}\notag
        ||z||_q &= sup\{z^Tx \enspace  ||x||_p \le 1\} \enspace and  \enspace \frac{1}{p} + \frac{1}{q}  =1
    \end{aligned}
\end{equation}

\textit{Grug:}
\begin{equation}
    \begin{aligned}\notag
        E(L^\sim ) &= E(\frac{1}{N}\sum_{t=0}^{N-1}L^\sim(f(F+\delta_t,M+\gamma_t,W),y))\\
        &\approx {\frac{1}{N}\sum_{t=0}^{N-1}L(f(F + \delta_t,M,W),y) + \frac{1}{N}\sum_{t=0}^{N-1}\frac{\gamma_t}{2}  \cdot \partial _M L(f(F+\delta_t,M,W),y)} \\
        &\approx {L(f(A ,M,W),y) + \frac{1}{N}\sum_{t=0}^{N-1}\frac{\delta_t}{2}  \cdot \partial _F L(f(F,M,W),y)} + \\
        &\frac{1}{N}\sum_{t=0}^{N-1}\frac{\gamma_t}{2} \cdot \partial _M (L(f(F ,M,W),y) + \frac{\delta_t}{2}  \cdot \partial _F L(f(F,M,W),y)) \\
        &= L + \frac{1}{N}\sum_{t=0}^{N-1}\frac{\delta_t}{2} \cdot \partial _F L + \frac{1}{N}\sum_{t=0}^{N-1}\frac{\gamma_t}{2} \cdot \partial _M L + \frac{1}{N}\sum_{t=0}^{N-1}\frac{\gamma_t}{2} \cdot \frac{\delta_t}{2} \cdot \partial _F (\partial _M L)\\
        & \approx L + \frac{1}{2N}\sum_{t=0}^{N-1}\varepsilon_t \cdot || \partial_F L ||_{q_t} + \frac{1}{2N}\sum_{t=0}^{N-1}\xi_t  \cdot || \partial_F L ||_{h_t} \\
    \end{aligned}
\end{equation}
where $\delta$ and $\gamma$ are the trained $F$ perturbation, which are defined as follows:

\begin{equation}
    \begin{aligned}\notag
                \delta_t &=\delta_{t-1} + \beta \cdot 
 \frac{\partial_{\delta} L(f(F+\delta_{t-1},M,W),y)}{||\partial_{\delta} L(f(F+\delta_{t-1},M,W),y)||}\\
                \gamma_t &=\gamma_{t-1} + \alpha \cdot 
 \frac{\partial_{\alpha} L(f(F+\gamma_{t-1},M,W),y)}{||\partial_{\gamma} L(f(F+\gamma_{t-1},M,W),y)||}
    \end{aligned}
\end{equation}

 $||\cdot||_{p_t}$ and $||\cdot||_{l_t}$ are the dual norm of $||\cdot||_{q_t}$ and $||\cdot||_{h_t}$,which are defined as follows:
\begin{equation}
    \begin{aligned}\notag
        ||z||_q &= sup\{z^Tx \enspace  ||x||_p \le 1\} \enspace and  \enspace \frac{1}{p} + \frac{1}{q}  =1\\
        ||z||_h &= sup\{z^Tx \enspace  ||x||_l \le 1\} \enspace and  \enspace \frac{1}{h} + \frac{1}{l}  =1
    \end{aligned}
\end{equation}

\textbf{Detailed proof of Theorem 2.}

\textbf{Theorem 2}~\textit{Grug can adjust the sample variance to achieve the stability of the training process .}

In order to make the conclusion more general, we assume the original $M$  is $1_{n\times n}$, which means that every element in $M$ is $1$.Thus, we can calculate its sample variance via the 1-norm of the $M$.

The sample variance of FLAG can be expressed as follows:
\begin{equation}
    \begin{aligned}\notag
        V(|FLAG|) &= V(M+\gamma_t) - V(M+\gamma_{t-1})\\
        &= V(\gamma_t - \gamma_{t-1})\\
        &= V(\alpha \cdot \textbf{A})\\
                &= \alpha^2 V(\textbf{A})\\
    \end{aligned}
\end{equation}
where $\textbf{A}$ is a distribution containing only 1 and -1, so the range of $A$ variance is from 0 to 1. So the range of FLAG variance can be calculated as follows:
$$0 \le V(FLAG) \le \alpha^2$$
Similarly, we can calculate range of \textit{Grug} variance as follows:
$$0 \le V(\textit{Grug}) \le \alpha^2 + \beta^2$$
DropMessage has been proven to have the smallest sample variance among random dropping methods, which can be expressed as follows:

\begin{equation}
    \begin{aligned}\
        V(|DropMessage|) &= V(M_\eta) - V(M_{\eta^\sim})\\
        &= V(M \times \textbf{R})\\
        &=V(M)\cdot V(\textbf{R}) + V(M) \cdot E(\textbf{R}) ^ 2 + V(\textbf{R}) \cdot E(M) ^ 2
    \end{aligned}
\end{equation}
where $\textbf{R}$ is a distribution containing 1, 0 and -1, so the range of its expectations is from 0 to 1 and the range of its variance is from 0 to 1 too. So we calculate the range of DropMessage as follows:
$$ 0 \le  V(|DropMessage|) \le 2V(M)  + E(M) ^2 $$

\textbf{Detailed proof of Theorem 3.}

\textbf{Theorem 3}~\textit{Grug perform regularization on 2 information dimension, namely feature and message, which is the optimal solution of existing regularization methods.} 

When we perform regularization on the adjacency matrix, the regularization term must contain $\partial_A L$. We can map this term to $\partial_M L$, which can be expressed as:
 $$ \partial_A L =  \frac{\partial L}{\partial M} \cdot \frac{\partial M }{\partial A} = \partial_M L  \cdot  \partial_A M
$$ 
where $\partial_A M$ is a function involving $\partial_F$.

Similarly, we can express regularization terms on the feature matrix as:
 $$ \partial_F L =  \frac{\partial L}{\partial M} \cdot \frac{\partial M }{\partial F} = \partial_M L  \cdot  \partial_F M
$$ 
where $\partial_f M$ is a function involving $\partial_A$.

For regularization $\partial_M L$ on the message matrix, there are two inferences as follows:
 $$ \partial_M L =  \frac{\partial L}{\partial F} \cdot \frac{\partial F }{\partial M} = \partial_F L  \cdot  \partial_M F
$$ 
 $$ \partial_M L =  \frac{\partial L}{\partial A} \cdot \frac{\partial A }{\partial M} = \partial_A L  \cdot  \partial_M A
$$ 
Here we take the regularization of node dimension as an example. When we only apply regularization on the message dimension, the regularization on the node dimension can be expressed as:
 $$ \partial_F L =  \frac{\partial L}{\partial M} \cdot \frac{\partial M }{\partial F} = \partial_M L  \cdot  \partial_F M
$$ 

\begin{figure}[htbp]
    \centering
    \includegraphics[scale=0.5]{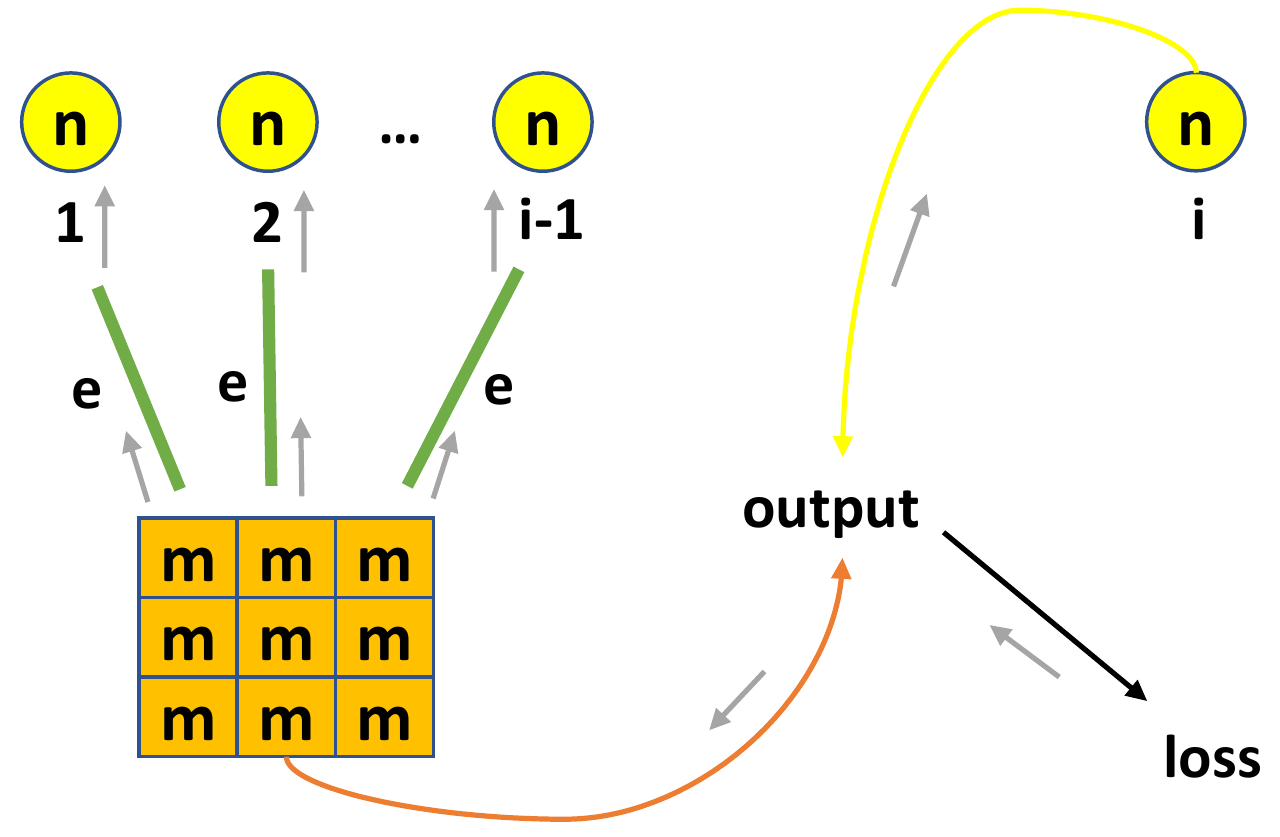}
    \caption{Process of gradient generation for node $i$.}
\label{gradient_generate}

    \end{figure} 

Figure \ref{gradient_generate} shows the process of gradient generation for node $i$. We find that regularization on dimensions is convertible, which means that when performing regularization on one dimension, it may also apply regularization in other dimensions at the same time (in one epoch). More specifically, the conversion is as follows:
\begin{itemize}
    \item Node: nodes.
    \item Message: message, edges and a part of nodes.
    \item Edge: edges and a part of nodes.
\end{itemize}

So in one epoch, in order to apply regularization on all dimensions, the optimal method is performing regularization on the message dimension and node dimension. In order to make our inferences interpretable, we take performing regularization on message dimension as an example.

\textbf{Definition 1}~\textit{For $\partial_F M_{i, j}$ in $\partial_F M$, it can be expressed as partial derivative of $M_j$ to $F_i$ and its value is 0 if $M_j$ is not generated by $F_i$}
$$
\begin{bmatrix}  
\partial_{x_1} m_1 \quad \partial_{x_2} m_1 \quad \dots \quad \partial_{x_p} m_1 \\  \partial_{x_1} m_2 \quad \partial_{x_2} m_2 \quad \dots \quad \partial_{x_p} m_2 \\  \dots \quad \dots \quad \dots \quad \dots \\  \partial_{x_1} m_k \quad \partial_{x_2} m_k \quad \dots \quad \partial_{x_p} m_k
\end{bmatrix}
$$

In the training process of node $n_i$ in one epoch $e_k$, node $n_i$ receives message $m_i$ from neighbor nodes ($n_1$, $n_2$ ... $n_{i-1}$). Performing regularization on message $m_i$, according to Chain rule (Rule of composed derivatives), is equivalent to performing regularization on $n_1$, $n_2$ ... $n_{i-1}$ without $n_i$ in a certain rule. However, in next epoch $e_{i+1}$, regularization on message $m_i$ can convert to node $n_i$, which can be expressed as follows:

\begin{equation}
    \begin{aligned}\notag
        M_{i+1} &= M_{i} - \psi\cdot reg (\partial_{M_{i}} L_{i})\\
    \end{aligned}
\end{equation}

\begin{equation}
    \begin{aligned}\notag
        F_{i+1} &= F_{i} - \psi \cdot \partial_{F_{i}} L_{i}\\
    \end{aligned}
\end{equation}

\begin{equation}
    \begin{aligned}\notag
        \partial_{F_{i+1}} L_{i+1} &= \frac{\partial ( (F_{i} - \psi \cdot \partial_{F_{i}} L_{i}) \cdot (M_{i} - \psi\cdot reg (\partial_{M_{i}} L_{i})) \cdot W )}{\partial (F_{i} - \psi \cdot \partial_{F_{i}} L_{i} )} \\
    \end{aligned}
\end{equation}

Despite regularization on $m_i$ can convert to $n_i$ in the training process with the epochs increasing. This conversion is noninterpretable and uncontrollable.

\textbf{Detailed proof of Theorem 4.}

\textbf{Theorem 4}~\textit{Message-passing HGNNs using Grug remain one and unique solution, which maintains parameters easily training.} 

In fact, the concavity and convexity of the function will directly affect the convergence rate of the parameters. Here we express $Grug$ (convex) as $F(x)$. So we can get the proportional relationship of the difference of the objective function between any 2 epochs $i$ and $*$.
\begin{equation}
    \begin{aligned}\notag
        F(x_i) - F(x_*)&\le F(x_{i-1}) - F(x_*) - \psi \cdot || \bigtriangledown F(x_{i-1}) ||^2 \\
    & \le F(x_{i-1}) - F(x_*) - \frac{\psi}{\omega} ( F(x_{i-1}) - F(x_*) ) \\
    &= (1- \frac{\psi}{\omega})(F(x_{i-1}) - F(x_*))
    \end{aligned}
\end{equation}
where $F(x_i) - F(x_{i-1}) \ge \omega || \bigtriangledown f(x_{i-1}) ||^2$ as $F(x)$ is convex. So when objective function $F(x)$ reach the least deviation $\vartheta$ in final epoch $T$, the inference is as follows:
\begin{equation}
    \begin{aligned}\notag
        F(x_T) - F(x_*)& \le \vartheta \\
        (1- \frac{\psi}{\omega})^T (F(x_{0}) - F(x_*)) &\le \vartheta \\
        \frac{1}{(1- \frac{\psi}{\omega})^T (F(x_{0}) - F(x_*))} &\ge \frac{1}{\vartheta}\\
        T &\ge \frac{\frac{(F(x_{0}) - F(x_*))}{\vartheta}}
        {\log \frac{1} {(1- \frac{\psi}{\omega})}}
    \end{aligned}
\end{equation}

So $T$ in $Grug$ function $F(x)$ is $O(\log \frac{1}{\vartheta})$ order.

For drop methods (non-convex), objective functions $F(x)$ is non-convex. When it converges, its first-order gradient satisfies as follows:
$$\bigtriangledown ||F(x_i)|| \le \vartheta $$
So in final epoch $T$, $F_{x_T}$ can be expressed:
\begin{equation}
    \begin{aligned}\notag
        F(x_T) &\le  F(x_0) - \psi \sum_{i=0}^{T-1} || \bigtriangledown F(x_{T-1}) ||^2\\
       \psi \sum_{i=0}^{T-1} || \bigtriangledown F(x_{T-1}) ||^2 &\le  F(x_0)- F(x_T) \le F(x_0) - F(x_*)\\
       \min || \bigtriangledown F(x_{T-1}) ||^2 &\le || \bigtriangledown F(x_{T-1}) ||^2 \le \frac{F(x_0) - F(x_*)}{\psi T}\\
       || \bigtriangledown F(x_{T-1}) ||^2 &\le O(\frac{1}{T}) \le \vartheta^2
    \end{aligned}
\end{equation}
So $T$ in drop method function $F(x)$ is $O(\frac{1}{\vartheta^2})$ and it is easy to find that $O(\frac{1}{\vartheta^2}) > O(\log \frac{1}{\vartheta})$. So compared to drop methods, $Grug$ needs fewer epochs to reach convergence.

\textbf{Detailed proof of Theorem 5.}

\textbf{Theorem 5}~\textit{Grug is able to maximize the utilization of diverse information in heterogeneous graph.} 

Since DropMessage has been proven to be the most capable of keeping diverse information in the drop method, we compare 3 methods: DropMessage, FLAG, \textit{Grug}.

1) More diverse information.

For each element $m_{i,j}$ in message matrix $M$, $m_{i,j}^\sim$ represents the information actually used in epoch $e_k$. In epoch $e_k$, $m_{i,j}^\sim$ can be expressed as follows:

DropMessage:
$$m_{i,j}^\sim = m_{i,j} \quad or \quad 0$$
FLAG/\textit{Grug}:
$$m_{i,j}^\sim = m_{i,j}  + \gamma_{i,j}^0 + \sum_{k=0}^{s-1} \psi \cdot \partial_{\gamma_{i,j}^k} L_k$$

where $r_0$ represents the initial value of $\gamma$, and $\psi$ represents the learning ratio for the training process. The $s$ is the total number of epochs, and
$k$ is the current training epoch. Here we use $L_k$ to represent the loss of the $kth$ epoch. The derivation of \textit{Grug} can be found in section \ref{Algorithm description}.

During the training process, the $L_k$ reduces gradually in each epoch, so the $\partial_{\gamma_{i,j}^k} L_k$ also decreases in each epoch. The value of $m_{i,j}^\sim$ is updated $s$ times, which means that $m_{i,j}^\sim$ does not generate duplicate values in the $s$ times. Therefore, the $m_{i,j}^\sim$ of FLAG and Grug is higher than the $m_{i,j}^\sim$ of DropMessage.

For the FLAG, it only performs regularization on the node dimension without considering the message dimension. This must result in generating duplicate values in $m_{i,j}^\sim$ because several messages may come from the same node. Here $Q()$ represents the amount of $m_{i,j}^\sim$ in the training process. So the $Q()$ of Grug, FLAG and DropMessage can be expressed as follows:
$$Q(Grug) > Q(FLAG) > Q(DropMessage)$$

2) Retain original information.

Additional experiments \ref{Additional Experiments} show that drop methods have the risk of over-fitting. We hypothesize that one of the causes of overfitting is the loss of original information. Here we theoretically analyze the ability to retain original information between the drop methods and \textit{Grug}. For the DropMessage, if the drop ratio is $\chi$, the information loss ratio of DropMessage is also $\chi$.

For FLAG and $Grug$, we assume that $m_{i, j}$ loses $d_{i,j}$ information in epoch $e_k$, which can be expressed as:
\begin{equation}
    \begin{aligned}\notag
    d_{i,j} &=  (m_{i,j} + \alpha \frac{\partial_{m_{i,j}} L}{|\partial_{m_{i,j}} L|} - m_{i,j} )/ m_{i,j}\\
    &= \alpha \frac{\partial_{m_{i,j}} L}{|\partial_{m_{i,j}} L|}) / m_{i,j}\\
    &\propto \partial_{m_{i,j}} L
    \end{aligned}
\end{equation}

The value of $d_{i,j}$ depends on $\partial_{m_{i,j}} L$. Besides, during the training processing, $d_{i,j}$ gradually decreases to 0 as gradient $\partial_{m_{i,j}} L$ gradually declines to 0. In summary, the ratio of loss information can be found as follows:
$$Grug, FLAG < DropMessage$$

\subsection{Experiment Details} \label{Experiment Details}

\begin{table}[htbp]
\centering
\begin{tabular}{cccc}
\hline
Dataset & Heterogeneous Nodes & Heterogeneous Edges & Features \\ \hline
ACM     & 5912/3025/57        & 9936/3025           & 1902     \\
DBLP    & 4057/14328/20/8789  & 19645/14328/88420   & 334      \\
IMDB    & 4661/5841/2270      & 13983/4661          & 1256     \\
amazon  & 10099               & 76924/71735         & 1156     \\
LastFM  & 1892/17632/1088     & 92834/25434/23523   & 0        \\ \hline
\end{tabular}
\vspace{2mm}
\caption{Dataset Statistics.}
\label{Dataset Statistics}
\end{table}

\textbf{Hardware specification and environment.} We run our experiments on the machine with 12th Gem Inter(R) Core(TM) (2.70 GHz), NVIDIA RTX 3050 GPUs (16 GB). The code is written in Python 3.9 and Pytorch 1.12.1.

\textbf{Dataset statistics.} Table \ref{Dataset Statistics} shows the statistics of datasets.

\begin{table}[htbp]
\centering
\begin{tabular}{ccc}
\hline
Dataset   & $\alpha$  & $\beta$   \\ \hline
RGCN-ACM  & 0.35   & 0.01   \\
RGAT-ACM  & 0.08   & 0.01   \\ \hline
RGCN-DBLP & 0.11   & 0.01   \\
RGAT-DBLP & 0.0001 & 0.001  \\ \hline
RGCN-IMDB & 0.002  & 0.0001 \\
RGAT-IMDB & 0.001  & 0.0004 \\ \hline
\end{tabular}
\vspace{2mm}
\caption{Statistics of Optimal $\alpha$ and $\beta$ for Node Classification.}
\label{op-nc}
\end{table}

\begin{wraptable}{i}{7cm}
\centering
\begin{tabular}{ccc}
\hline
Dataset   & $\alpha$  & $\beta$   \\ \hline
RGCN-amazon  & 0.0002  & 0.0001  \\
RGAT-amazon  & 0.003   & 0.001   \\ \hline
RGCN-LastFM & 0.002   & 0.001   \\
RGAT-LastFM & 0.001 & 0.002  \\ \hline
\end{tabular}
\vspace{2mm}
\caption{Statistics of Optimal $\alpha$ and $\beta$ for Link prediction.}
\vspace{-1mm}
\label{op-lp}
\end{wraptable}

\textbf{Implementation details.} We conduct 10 independent experiments for each setting and obtain the average results. We apply 1-layer models to the three datasets of node prediction tasks (ACM, DBLP, IMDB). However, for the datasets of link prediction tasks (amazon, LastFM), we use three-layer models. In all cases, we use the Adam optimizer with learning rate of 0.001 and train each model 200 epochs. We adjust the $\alpha$ and $\beta$ from 0.0001 to 0.5 in steps of 0.0001 and select the optimal one for each setting. Table \ref{op-nc} and \ref{op-lp} present the optimal parameter we selected.

\subsection{Additional Experiments} \label{Additional Experiments}

In section \ref{stability}, we present that \textit{Grug} can directly control the sample variance to maintain the stability of the training process. We conduct experiments and record training loss and validation loss fluctuation throughout the whole training process. Figure \ref{fig: Loss} shows the change of loss in RGCN training processes when employing different gradient regularization methods on the ACM dataset.  We have the following observations:

\begin{figure}[!h]
\centering
 \subfigure[Training Process]{
 \includegraphics[scale=0.41]{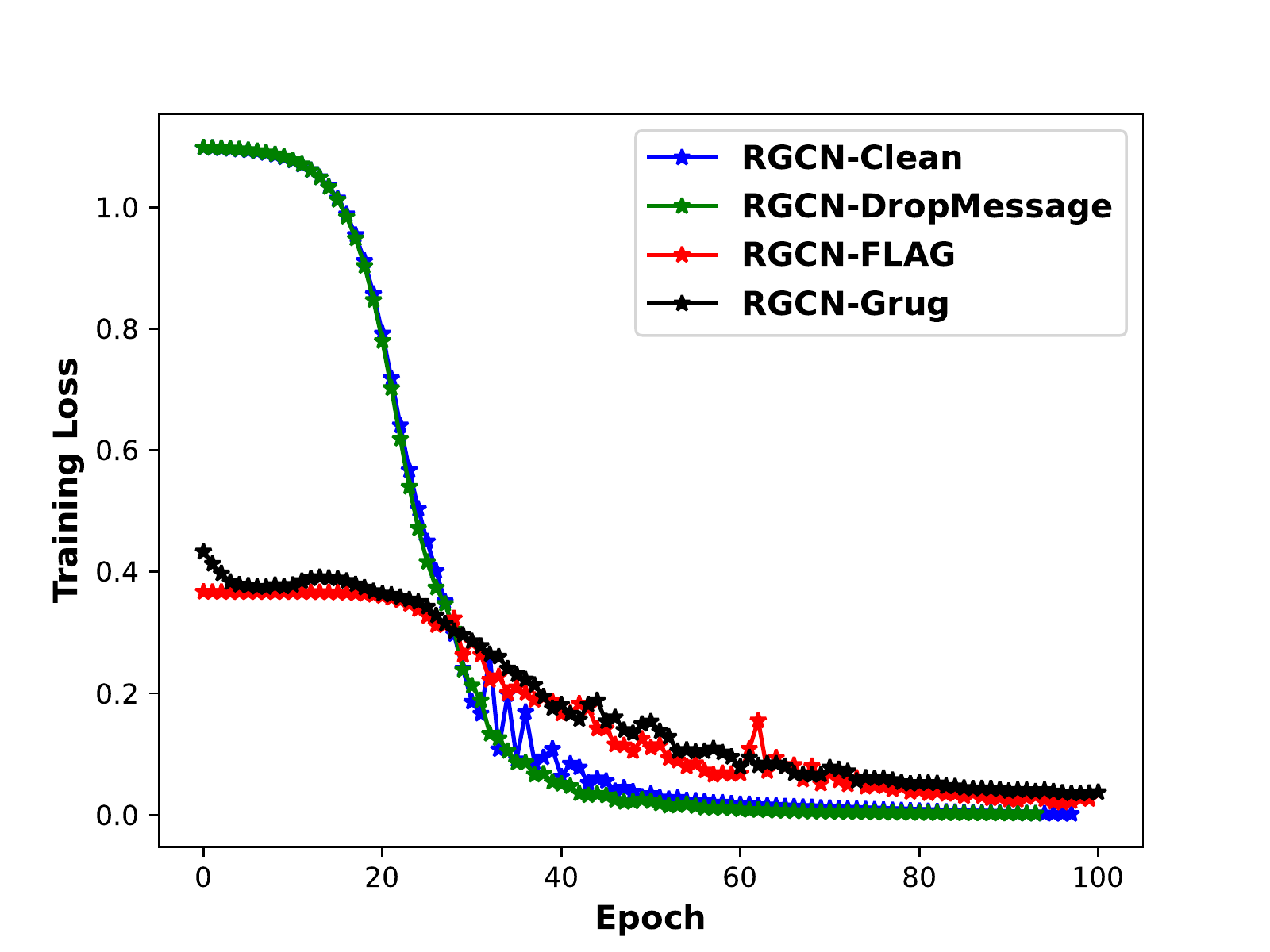}
   \label{fig: training Loss} 
    }
 \subfigure[Validation Process]{
 \includegraphics[scale=0.41]{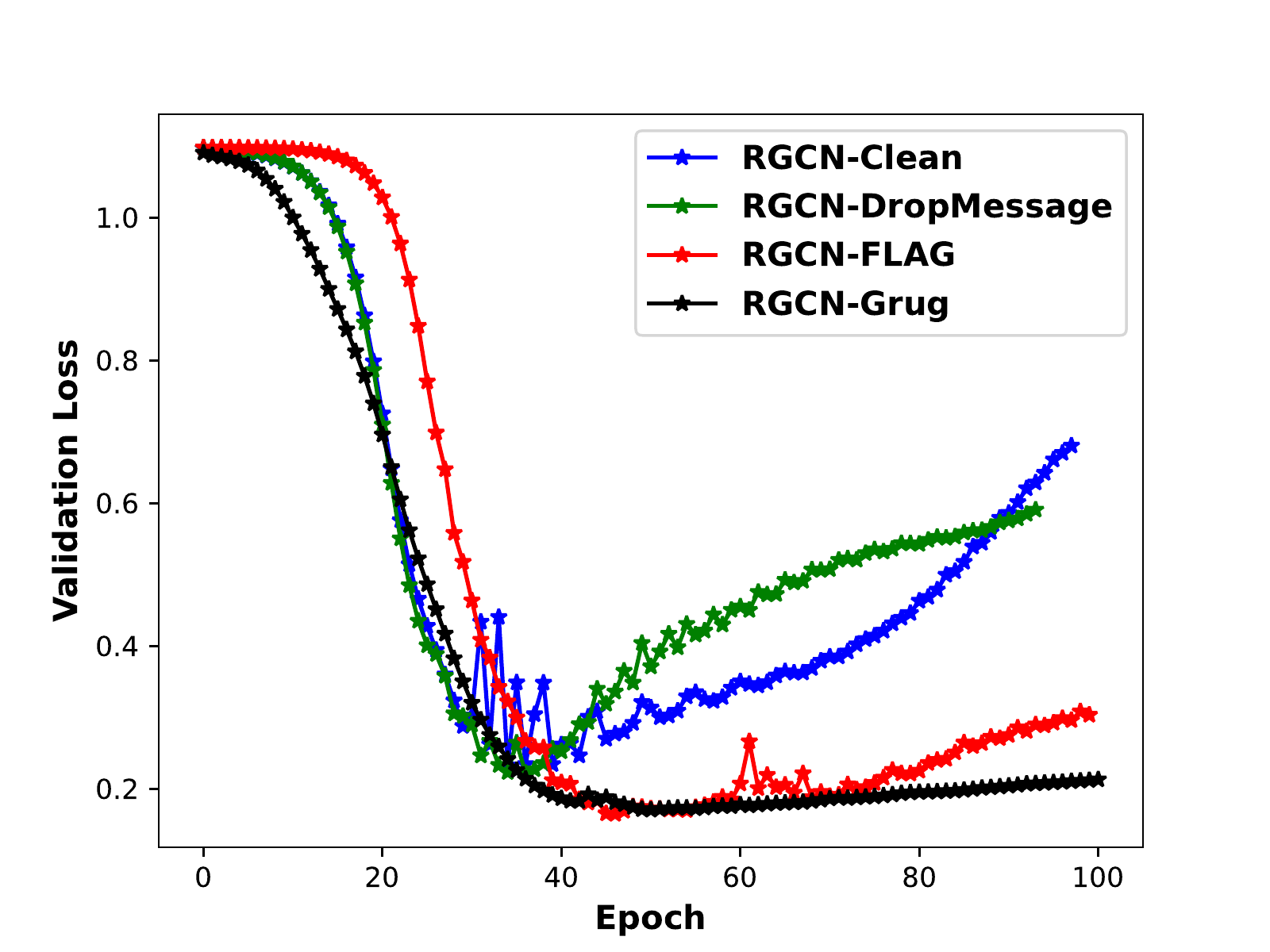}
   \label{fig: val Loss} 
    }
\caption{Training Process and Validation Process Analysis.}
  \label{fig: Loss} 
\end{figure}

\textbf{Training process analysis.} 

1) In training process, loss of \textit{Grug} declines more stability than DropMessage. 

2) In the validation process, the performance of \textit{Grug} keeps stable while other methods suffer from over-fitting.

3) DropMessage quickly reach convergence (around epoch 50), but the problem of over-fitting for it compares to \textit{Grug}.

Observation 1 presents the stability of \textit{Grug}, which is consistent with the theoretical results in Section \ref{stability}. Observation 2 suggests that \textit{Grug} receives more diverse information than other methods. Observation 3 indicates that \textit{Grug} keeps the original information while DropMessage is not. This verifies our theory in section \ref{diversity}. In conclusion, taking \textit{Grug} as the gradient regularization method can obtain more diverse information and stability in the face of over-fitting problems.

\textbf{Ablation experiments}

The ablation studies on all the five datasets are summarized in Table \ref{Ablation}.
We conduct 4 ablation regularization methods as a comparison.
\begin{itemize}
    \item $Grug_n:$ only performs regularization on the node dimension.
    \item $Grug_e:$ only performs regularization on the edge dimension.
    \item $Grug_m:$ only performs regularization on the message dimension.
    \item $Grug_T:$ performs regularization on the message dimension, node dimension and edge dimension.
\end{itemize}

\begin{wraptable}{i}{9cm}
\centering
\begin{tabular}{ccccccc}
\hline
\multicolumn{2}{c}{Task}        & \multicolumn{3}{c}{Node Classification}          & \multicolumn{2}{c}{Link prediction} \\
\multicolumn{2}{c}{Dataset}     & ACM            & DBLP           & IMDB           & amazon           & LastFM           \\ \hline
\multirow{6}{*}{\rotatebox{90}{\textbf{RGCN}}} & $Grug_n$ & 92.43          & 94.22          & 56.74          & 78.87            & 56.96            \\
                      & $Grug_e$ & 91.08          & 93.97          & 56.55          & 77.32            & 55.92            \\
                      & $Grug_m$ & 91.22          & 94.47          & 56.99          & 78.56            & 56.89            \\ \cline{2-7} 
                      & $Grug_T$ & 93.30          & 95.05          & 60.05          & 80.99            & 60.56            \\
                      & $Grug$    & 93.27          & 95.08          & 59.92          & 80.93            & 60.37            \\
                      & $GAP$     & \textbf{-0.03} & \textbf{+0.03} & \textbf{-0.13} & \textbf{-0.06}   & \textbf{-0.19}   \\ \hline
\multirow{6}{*}{\rotatebox{90}{\textbf{RGAT}}} & $Grug_n$ & 91.46          & 93.38          & 56.09          & 85.59            & 78.66            \\
                      & $Grug_e$ & 90.56          & 93.45          & 56.98          & 83.11            & 77.65            \\
                      & $Grug_m$ & 90.97          & 93.94          & 57.37          & 84.97            & 78.95            \\ \cline{2-7} 
                      & $Grug_T$ & 92.90          & 94.01          & 57.88          & 86.67            & 80.37            \\
                      & $Grug$    & 92.84          & 94.21          & 57.79          & 86.84            & 80.39            \\
                      & $GAP$     & \textbf{-0.06} & \textbf{+0.20} & \textbf{-0.09} & \textbf{+0.17}   & \textbf{+0.02}   \\ \hline
\end{tabular}
\vspace{2mm}
\caption{Ablation results.}
\label{Ablation}
\end{wraptable}

In section \ref{universality}, we suggest performing regularization on the message dimension and node dimension as a selection without considering on edge dimension. Table \ref{Ablation} shows the gap between $Grug_T$ and $Grug$. The average gap is only 0.013 for node classification and 0.015 for link prediction. Besides, performing regularization on edge dimensions needs to generate a new adjacency matrix in each epoch, which consumes extra time and memory. So \textit{Grug} choose message dimension and node dimension as a substitute for all dimensions. On the other hand, in 10 settings, $Grug_m$ and $Grug_n$ achieve the optimal results in 5 settings and 5 settings respectively. However, $Grug_e$ performs not well as the other 2 methods. In conclusion, \textit{Grug} performing regularization on the message and node dimension is the optimal strategy.

\textbf{Gradient Norm Analysis.}
\begin{figure}[!h]
\centering
 \subfigure[Euclidean norm (L2 norm)]{
 \includegraphics[scale=0.41]{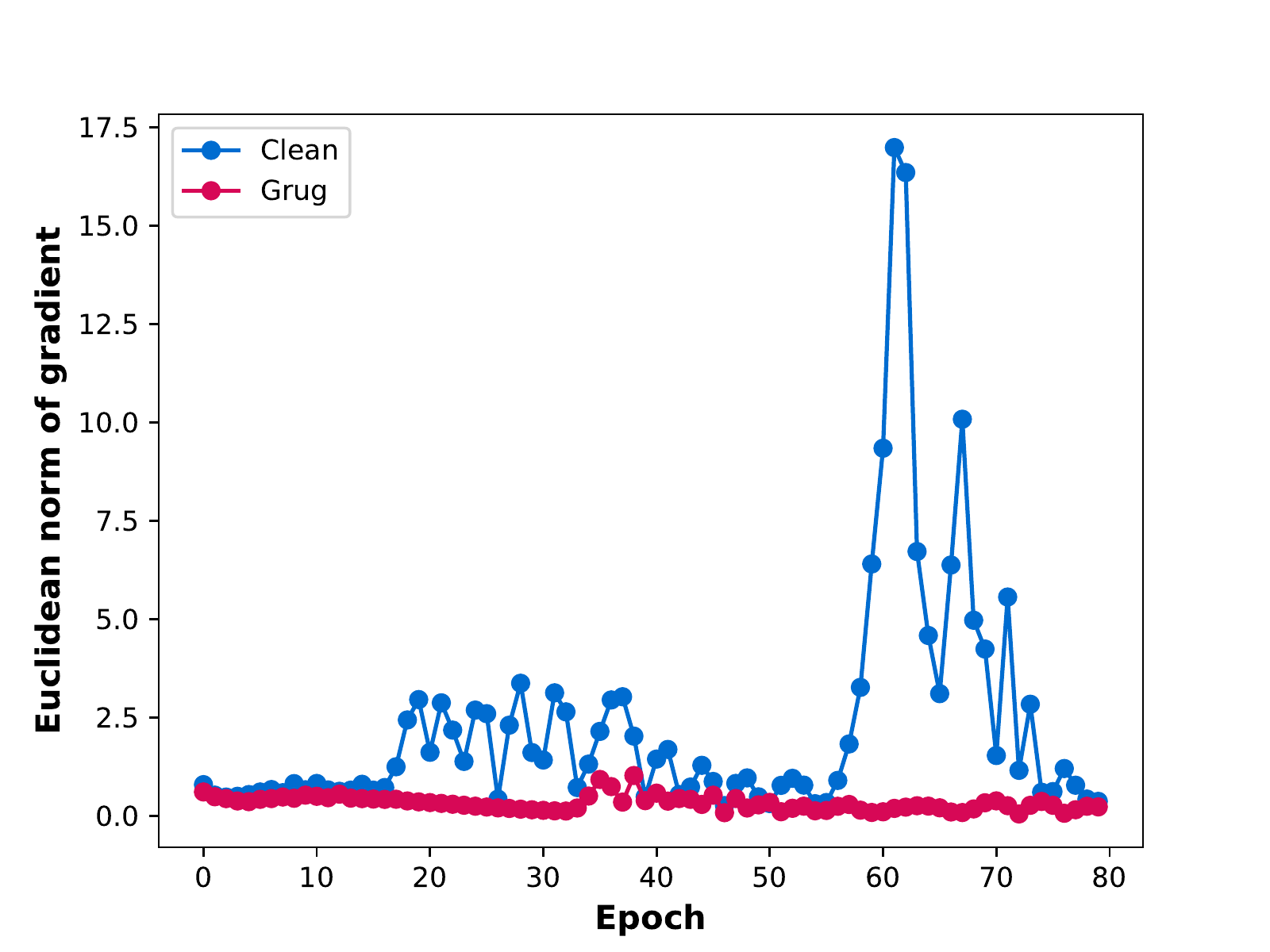}
   \label{fig: L2_0} 
    }
 \subfigure[Manhattan norm (L1 norm)]{
 \includegraphics[scale=0.41]{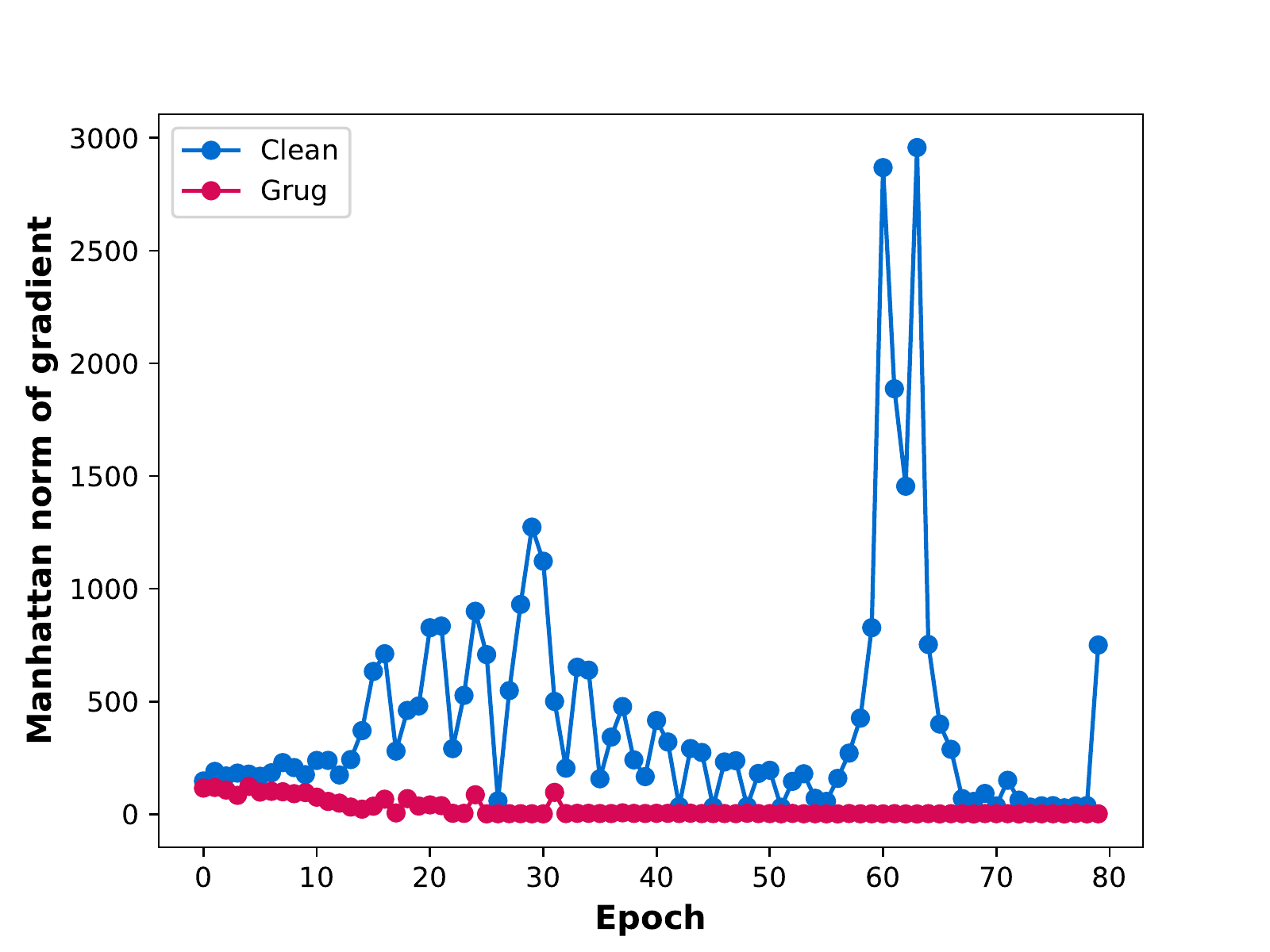}
   \label{fig: L2_1} 
    }
\caption{Paradigm norm of gradient analysis.}
  \label{fig: Paradigm} 
\end{figure}

As we mentioned in section \ref{simplicity}, the 1-dimension Paradigm norm and 2-dimension Paradigm norm can be regarded as a specific form of \textit{Grug}. Figure \ref{fig: Paradigm} shows the Euclidean norm and Manhattan norm of the gradient in the training process. We find that using \textit{Grug} can make the  1-dimension Paradigm and 2-dimension Paradigm small in the whole training process. Figure \ref{fig: L2_0} shows the Euclidean norm between model and mode using \textit{Grug}, suggesting that \textit{Grug} performs better in narrowing the 2-dimension Paradigm of the gradient. Besides Figure \ref{fig: L2_1} indicates that \textit{Grug} can also limit the value of the 1-dimension Paradigm of the gradient. Therefore, compared with previous methods, the gradient of \textit{Grug} can remain stable in the training process without great fluctuation. A smoother gradient generated by \textit{Grug} can significantly improve the speed and efficiency of training convergence of the model. This experimental result is consistent with the theoretical proof of section \ref{simplicity}.








\end{document}